# From Drawings to Decisions: A Hybrid Vision-Language Framework for Parsing 2D Engineering Drawings into Structured Manufacturing Knowledge

**Muhammad Tayyab Khan [a, c*], Lequn Chen [b*], Zane Yong [b], Jun Ming Tan [b], Wenhe Feng [a], Seung Ki Moon [c*]**

[a] Singapore Institute of Manufacturing Technology (SIMTech), Agency for Science, Technology and Research (A*STAR), 5 CleanTech Loop, #01-01 CleanTech Two Block B, Singapore 636732, Republic of Singapore

[b] Advanced Remanufacturing and Technology Centre (ARTC), Agency for Science, Technology and Research (A*STAR), 3 CleanTech Loop, #01-01 CleanTech Two, Singapore 637143, Republic of Singapore

[c] School of Mechanical and Aerospace Engineering, Nanyang Technological University, 639798, Singapore

[*] Corresponding authors: khan0022@e.ntu.edu.sg (M.T. Khan), chen1470@e.ntu.edu.sg (L. Chen), skmoon@ntu.edu.sg (S.K. Moon)

## Abstract

Efficient and accurate extraction of key information from 2D engineering drawings is essential for advancing digital manufacturing workflows. This information includes elements such as geometric dimensioning and tolerancing (GD&T), measures, material specifications, and textual annotations. Manual extraction remains slow and labor-intensive, while generic optical character recognition (OCR) models often fail to interpret 2D drawings accurately due to complex layouts, engineering symbols, and rotated annotations. These limitations result in incomplete and unreliable outputs. To address these challenges, this paper proposes a hybrid vision-language framework that integrates a rotation-aware object detection model (YOLOv11-obb) with a transformer-based vision-language parser. We introduce a structured parsing pipeline that first applies YOLOv11-obb to localize annotations and extract oriented bounding box (OBB) image patches, which are subsequently parsed into structured outputs using a fine-tuned, lightweight vision-language model (VLM). To develop and evaluate this pipeline, we curate a dataset of 1,367 2D mechanical drawings manually annotated across nine key categories: GD&Ts, General Tolerances, Measures, Materials, Notes, Radii, Surface Roughness, Threads, and Title Blocks. YOLOv11-obb is trained on this dataset to detect OBBs and extract annotation patches. These image patches are then parsed using two fine-tuned open-source VLMs. The first is Donut, a transformer-based model that combines a Swin-B visual encoder with a BART text decoder, enabling end-to-end parsing directly from images without relying on OCR. The second is Florence-2, a prompt-driven encoder-decoder model that integrates a DaViT vision backbone and supports structured output generation through multimodal token alignment. Both models are lightweight and well-suited for specialized industrial tasks under limited computational overhead. Following fine-tuning of both models on the curated dataset of image patches paired with structured annotation labels, a comparative experiment is conducted to evaluate parsing performance across four key metrics. Donut outperforms Florence-2, achieving 89.2% precision, 99.2% recall, and a 94% F1-score, with a hallucination rate of 10.8%. Finally, a case study demonstrates how the extracted structured information supports downstream manufacturing tasks such as process and tool selection, showcasing the practical utility of the proposed framework in modernizing 2D drawing interpretation.

## 1. Introduction

Engineering drawings remain fundamental to manufacturing, conveying essential information such as geometric dimensions, tolerances, surface finishes, and annotations that directly affect product quality, production efficiency, and cost [1]. Two-dimensional (2D) drawings serve as a critical communication interface between design and manufacturing, guiding downstream tasks, including tool selection, process planning, inspection, and quality assurance. Among the most important annotations are Geometric Dimensioning and Tolerancing (GD&T) symbols, which are standardized under ASME Y14.5-2018 and encode design intent and permissible geometric variation [2]. Accurate extraction of the drawing information is vital for ensuring production quality and cost-effectiveness, as misinterpretations or extraction errors can lead to defective components, incorrect machine setups, and costly delays [3].

Despite the availability of digital tools, interpreting 2D engineering drawings remains a labor-intensive and slow task. This is especially true for complex drawings with dense annotations or intricate GD&T callouts. Manual practices such as ballooning, where features and tolerances are transcribed into spreadsheets or inspection reports, remain common even in digital formats. Semi-automated tools like AutoCAD's ballooning utility [4] and commercial platforms such as Mitutoyo's MeasurLink [5] offer partial assistance but rely heavily on human input, limiting scalability and introducing inconsistency. Automating information extraction from engineering drawings holds significant potential for improving efficiency, consistency, and digital integration.

Recent advances in deep learning (DL) have aimed to reduce manual effort. Object detection models, particularly those based on the *You Only Look Once* (YOLO) architecture [6], have been employed to localize regions of interest in drawings, while optical character recognition (OCR) tools have been used to extract text. However, these models often struggle with real-world engineering drawings, which include rotated symbols, stylized fonts, and complex layouts. Generic OCR models typically produce segmentation errors and unstructured outputs that require extensive post-processing [7].

To address these challenges, some recent studies have explored hybrid pipelines that combine computer vision techniques with vision-language models (VLMs) to enhance the interpretation of technical documents. For example, the eDOCr2 framework [8] segments engineering drawings into functional zones such as title blocks, dimensions, and feature control frames (FCFs), and uses general-purpose VLMs like Qwen2-VL-7B and GPT-4o for semantic analysis. While promising for design validation, these systems rely on models that are not fine-tuned for engineering contexts, and therefore often produce hallucinated or inaccurate outputs when parsing domain-specific symbols and complex layouts. Moreover, our previous work employs a fine-tuned VLM pipeline specifically tailored for GD&T extraction from 2D engineering drawings [7]. By adapting to the visual and symbolic characteristics of these documents, improved accuracy over generic models is achieved. However, the pipeline processes entire drawings in a single pass, which results in performance degradation for densely annotated inputs. As the number of annotations increases, the model struggles to resolve individual symbols and structures, often leading to missed or incorrect elements. These limitations underscore the need for more localized, modular, and domain-adapted methods for interpreting engineering drawing.

The automated interpretation of engineering drawings continues to face two persistent challenges. First, accurately localizing diverse annotation types requires computer vision models capable of handling variations in layout, orientation, and scale. Second, parsing these annotations requires models fine-tuned to the visual and symbolic conventions of engineering documentation. Generic models often fail to extract structured knowledge due to symbol misclassification, inconsistent formatting, or lack of domain-specific adaptation.

To address these challenges, this paper proposes a novel and hybrid vision-language framework for structured information extraction from 2D engineering drawings. The system follows a two-stage architecture. In the first stage, YOLOv11-obb [9] is used to localize and extract annotation regions across the drawing space. In the second stage, two open-source VLMs, Donut [10] and Florence-2 [11], are fine-tuned on a curated dataset of annotation image patches paired with structured ground truths. These fine-tuned models are then used to semantically parse the localized regions and generate structured outputs. Donut and Florence-2 are selected for their lightweight architecture and suitability for task-specific adaptation in constrained environments. A comparative analysis is conducted to assess parsing performance against manually verified ground truth using a structured test set. Evaluation is carried out using four key metrics: precision, recall, F1-score, and hallucination rate. This framework supports robust, category-aware parsing across diverse annotation types, and forms the basis for downstream integration into digital manufacturing workflows.

The main contribution of this work is a two-stage hybrid framework that combines rotation-aware object detection with fine-tuned vision-language parsing for extracting structured information from 2D drawings. To support this framework, a curated dataset of 1,367 annotated 2D drawings is used, and a comparative evaluation of two fine-tuned VLMs is conducted against manually verified ground truth. A case study illustrates the practical utility of the extracted structured outputs in downstream manufacturing tasks.

The remainder of this paper is organized as follows: Section 2 reviews related work on engineering drawing information extraction and concludes with identified research gaps. Section 3 details the proposed methodology, including dataset curation, annotation detection using oriented bounding boxes (OBBs), and VLM fine-tuning. Section 4 presents experimental results, including detection and parsing performance, as well as a qualitative validation. Section 5 provides a case study demonstrating downstream integration into digital manufacturing workflows. Finally, Section 6 concludes the paper and discusses directions for future research.

## 2. Literature Review

Advances in interpreting engineering drawing requires a comprehensive understanding of prior approaches, their limitations, and recent technological developments. This section reviews key contributions across three thematic areas: the evolution of drawing interpretation methods, the emergence of transformer-based models for structured understanding, and the integration of extracted information into digital manufacturing systems.

### 2.1 Traditional and Deep Learning-Based Annotation Extraction

Engineering drawings have long served as a primary medium for communicating design intent. While modern practices such as model-based definition (MBD) [12] and model-based systems engineering (MBSE) [13] advocate the use of enriched 3D models, many industries continue to rely heavily on conventional 2D drawings. These documents are frequently encountered as scanned images, PDFs, or rasterized CAD exports [14]. Such formats arise not only from historical archiving but also from deliberate dissemination strategies intended to protect intellectual property by reducing semantic leakage. Importantly, 2D engineering drawings, often CAD-derived and current rather than legacy, remain the primary carriers of manufacturing semantics including GD&T, tolerances, surface finish, and notes. While 3D CAD models capture geometry, they typically lack these annotations unless enriched through full MBD. Symbolic and spatial information in drawings remain essential for ensuring part functionality, product quality, and manufacturability. The importance of extracting and integrating this information into downstream manufacturing processes has long been recognized. For example, Gao et al. [15] introduced a framework to translate design tolerances into machining tolerances for computer-aided process planning (CAPP), thereby aligning GD&T specifications with manufacturing features. Sun and Gao [3] later extended this work by proposing a rule-based, datum-centric schema to enable consistent, machine-interpretable GD&T representation.

Initial efforts in automated interpretation relied predominantly on rule-based systems that decomposed drawings into orthographic views and extracted geometric features [16]. While these methods validated the feasibility of automation, they often fail under noisy or complex conditions. The emergence of DL provided an alternative in the form of data-driven models capable of learning patterns from large, annotated datasets. For instance, Xie et al. [17] proposed a pipeline that integrates a convolutional neural network (CNN) for region detection with a graph neural network (GNN) for structural reasoning. Their system achieved 97% precision in region detection and 90.8% accuracy in manufacturing method prediction, illustrating DL's potential to bridge CAD and CAM systems.

A critical aspect of engineering drawing interpretation is the extraction of annotations that complement geometric features. These include dimensions, notes, and standard symbols, all of which convey essential manufacturing semantics. However, this task remains challenging due to irregular text placement, varied font styles, overlaps with graphical elements, and domain-specific notation. Traditional OCR and template-matching techniques often fail under such conditions, driving the development of more robust alternatives.

Recent studies have increasingly adopted hybrid pipelines that combine image preprocessing, object detection, and OCR. For instance, Xu et al. [18] applied image noise filtering, block segmentation, and CNN-based character recognition to extract geometric tolerance specification callout (GTSC). Lin et al. [1] used YOLOv7 to detect drawing elements and employed Tesseract for text extraction, achieving up to 85% accuracy on industrial scans. Jamieson et al. [17] improved robustness by handling rotated and skewed text via CNN-based preprocessing. Francois et al. [19] incorporated a domain-specific post-OCR correction module, yielding 87.2% detection accuracy and 79.2% recognition accuracy. Khallouli et al. [14] developed a transformer-based OCR tailored to legacy shipbuilding drawings, which outperformed generic OCR systems in specialized domains. Beyond textual elements, engineering drawings include symbolic annotations such as surface finishes and FCFs, particularly within the GD&T framework. These visual entities are typically extracted using object detection techniques. CNN-based YOLO detectors have gained prominence due to their robustness in handling dense and noisy layouts. For instance, Mani et al. [20] developed specialized detectors that not only identify symbols but also associate them with adjacent text labels. Yu et al. [21] introduced a multi-detector framework for extracting symbols, texts, lines, and tabular legends from Piping and Instrumentation Diagrams (P&IDs).

Despite these advancements, significant challenges persist. Studies evaluating YOLO on heterogeneous engineering drawings report high precision for simple elements (87.6%) but a lower mean average precision (mAP) of 61% at an Intersection over Union (IoU) threshold of 0.5 [22]. Performance deteriorates in the presence of overlapping annotations, rare symbols and visually cluttered layouts. OCR systems continue to struggle with engineering-specific fonts, nested tolerance structures, and non-standard notation formats. Commercial tools such as Mitutoyo's MeasurLink and HighQA's Inspection Manager [23] offer some automation but rely on clean CAD inputs or standardized templates, limiting generalizability across diverse drawing types.

To address these limitations, recent work has explored data augmentation techniques and symbol-specific detection strategies to improve model generalization under class imbalance and visual variability [24]. These foundational advances have laid the groundwork for more sophisticated multimodal approaches that jointly leverage visual and textual features. Nevertheless, fully automated and robust parsing of engineering annotations remains an open challenge, particularly in domain-specific scenarios characterized by complex layouts and high annotation density.

## 2.2 Transformer-Based Models for Engineering Drawings

To address the limitations of traditional OCR and object detection pipelines, recent research has explored transformer-based models for structured understanding of engineering drawings. These models are capable of processing both visual and textual inputs [25], enabling integrated, context-aware document interpretation. In general document analysis, multimodal transformers such as LayoutLM [26] and DocFormer [27] have shown strong performance by jointly modeling text content and visual layout. Inspired by their success in natural document processing, researchers have begun adapting these architectures to technical domains, including engineering and schematic drawings. For example, Gu et al. [28] introduced ViRED, a transformer-based model designed to align graphical elements in circuit diagrams with corresponding entries in tabular annotations. The model achieved 96% accuracy in mapping components to the correct table entries, demonstrating the effectiveness of attention mechanisms in capturing visual-textual relationships in structured documents. Toro and Tarkian [8] proposed the eDOCr2 framework, segmenting engineering drawings into functional regions such as title blocks and FCFs, applies OCR, and subsequently prompts VLMs such as GPT-4o and Qwen2-VL-7B for semantic interpretation. This hybrid approach improves the quality of extracted information by using contextual inference to correct OCR outputs and impose structure on noisy scanned data. However, these systems are typically deployed in zero-shot settings, lack domain-specific alignment, and often produce hallucinated outputs or misinterpret engineering symbols. Additionally, their reliance on proprietary models limits customizability due to privacy, cost, and fine-tuning restrictions.

To address these limitations, some studies have explored fine-tuning open-source VLMs for engineering applications. In our prior work [7], Florence-2 was fine-tuned using 400 annotated mechanical drawings and demonstrated substantially higher precision and recall than closed-source models such as GPT-4o and Claude-3.5-Sonnet in zero-shot evaluations. These developments reflect a broader shift from fragmented, task-specific pipelines toward holistic, end-to-end models that treat engineering drawings as unified, multimodal documents. Transformer-based models, particularly those with vision-language capabilities, can represent spatial hierarchies, symbolic relationships, and semantic structures within a unified architecture. While their use in technical domains is still emerging, early results suggest strong potential for enhancing the reliability and scalability of engineering drawing interpretation, especially in applications requiring integration with downstream digital manufacturing systems.

### 2.3 Structured Drawing Information for Knowledge-Driven Manufacturing

Structured information extracted from 2D engineering drawings plays a vital role in downstream manufacturing tasks such as tool selection, process planning, and quality control. GD&T annotations, in particular, influence machining decisions and inspection strategies, where misinterpretation can result in non-conforming parts and costly rework [7].

Several studies have demonstrated the utility of structured drawing features for automation. Xie et al. [29] used them for classifying parts by manufacturing method, while Gao et al. [30] emphasized that missing tolerances hinder process planning. In the process industry, Dzhusupova et al. [31] showed how symbol pattern detection supports design validation. Such structured annotations also enable querying of datums, tolerance hierarchies, and material specifications, supporting traceable and intelligent decision-making. These applications align with knowledge-based engineering (KBE), which focuses on reusing and automating engineering knowledge [32].

When converted into structured formats; drawing content can populate ontologies and rule-based systems for downstream reasoning and control [33]. While MBE frameworks often focus on annotated 3D models, 2D drawings remain prevalent in practice[12], reinforcing the need for methods that convert them into interoperable digital formats. In this paper, process and tool selection is demonstrated as a downstream use case, using the structured output generated by the proposed hybrid extraction framework.

### 2.4 Research Gaps

The preceding literature review highlights substantial progress in automating information extraction from 2D engineering drawings using DL-based object detectors, OCR systems, and VLMs. These approaches have demonstrated potential for parsing both textual and symbolic annotations, and hybrid methods that combine object detection with VLMs have improved robustness across diverse layouts and annotation styles. However, several key limitations remain. Most existing solutions lack a domain-adapted, end-to-end framework to handle the complexity and variability of engineering drawings found in industrial practice. General-purpose VLMs are typically applied in zero-shot settings and are not fine-tuned on engineering-specific data, leading to frequent hallucination or misinterpretation of domain-specific graphical symbols. Additionally, the scarcity of publicly available, richly annotated datasets tailored to engineering drawings has constrained the development of models with broad generalization across annotation categories and visual styles. While some studies target isolated tasks such as symbol detection or textual parsing, few provide a comprehensive pipeline to transform raw drawing content into structured formats suitable for downstream manufacturing tasks. This gap limits the adoption of automated drawing interpretation in real-world production workflows. To address these challenges, this paper proposes a hybrid, domain-adapted framework that integrates rotation-aware object detection with a fine-tuned VLM for structured annotation extraction. Unlike prior work, the proposed approach emphasizes both annotation-level accuracy and downstream applicability, enabling integration into knowledge-driven manufacturing workflows.

## 3. Methodology

The proposed framework adopts a two-stage hybrid vision-language framework for structured information extraction from 2D engineering drawings, as illustrated in Fig. 1. In the first stage, a YOLOv11-obb model is trained to detect rotated and variably scaled annotation regions using OBBs. The trained model is then used to localize annotation patches across the full drawing set. These patches are paired with structured labels in JSON format to create a dataset for vision-language parsing. In the second stage, two open-source VLMs are fine-tuned to generate structured outputs from individual annotation patches, producing machine-readable semantic content. During inference, new drawings are processed through the trained YOLOv11-obb model to extract annotation regions, which are then parsed independently by the fine-tuned models. The resulting structured outputs are evaluated against manually verified ground truth using four metrics: precision, recall, F1-score, and hallucination rate. These outputs can be directly integrated into downstream applications. A case demonstration highlights how the extracted information supports tool and process selection in digital manufacturing workflows.

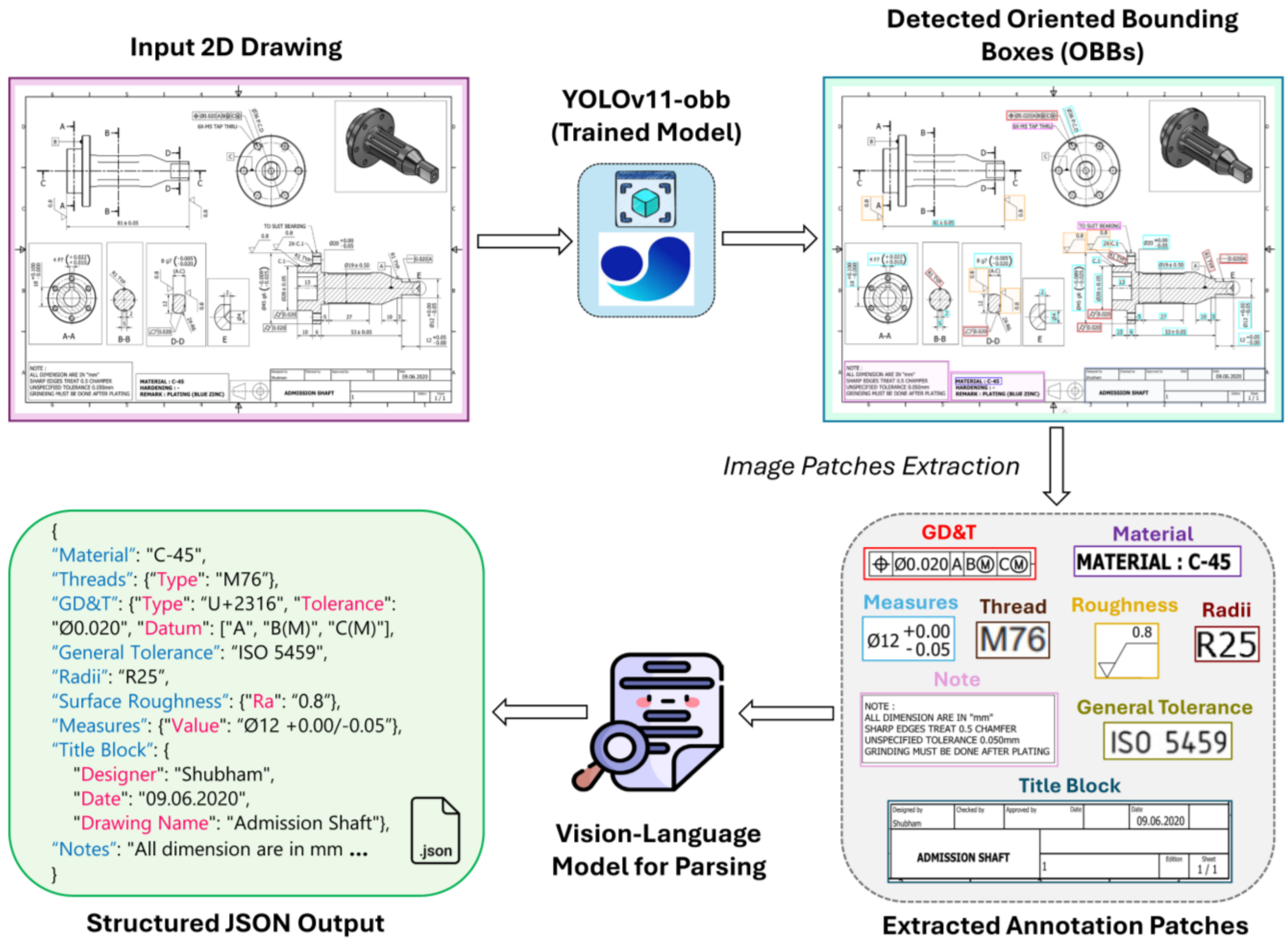

**Fig. 1.** Proposed two-stage hybrid vision-language framework for structured information extraction from 2D engineering drawings. The input drawing is first processed by YOLOv11-obb to detect OBBs. Detected OBBs are cropped into image patches representing individual annotation types. These patches are parsed by a fine-tuned VLM to generate structured outputs in JSON format, enabling downstream applications such as process and tool selection.

### 3.1 Dataset and Annotation Categories

To train and evaluate the proposed method, a new Engineering Drawing Annotation Dataset is curated, comprising 1,367 2D mechanical drawings collected from public datasets, standards documents, and open-source CAD repositories. The dataset includes a wide variety of drawing types, ranging from machined parts to assemblies, and spanning formats from clean CAD exports to scanned legacy blueprints, thereby reflecting the spectrum of documentation encountered in industrial practice. The collection covers multiple domains including aerospace, automotive, and general mechanical engineering, ensuring representativeness across diverse drafting conventions and complexity levels. All drawings are standardized to PNG format regardless of the original file type (e.g., PDF, JPEG). Each drawing is manually annotated using the Computer Vision Annotation Tool (CVAT) [34], with tight oriented bounding boxes applied to elements across nine key categories, ensuring consistent ground truth for benchmarking. The annotation categories are as follows:

- **GD&T:** Rectangular FCFs containing geometric symbols, tolerances, and datum references
- **General Tolerances**: Notes or tables specifying default tolerance values
- **Material Specifications**: Textual indicators of material type or treatment
- **Measures:** Linear and angular dimensional callouts
- **Notes:** General instructions or supplementary design details
- **Radii**: Radius-specific dimensional indicators
- **Surface Roughness**: Symbols denoting finish or texture requirements
- **Thread Callouts**: Designations for threaded features
- **Title Block:** Structured metadata, typically located in the bottom-right corner

These categories represent the most common and manufacturing-relevant annotation types observed across 2D drawings. It is important to note that this taxonomy is a practical grouping selected for model training and evaluation. It complements, but does not replace, established conventions that distinguish views from annotations and define stricter semantic rules for dimensional interpretation. The categorization reflects dataset-specific patterns while retaining alignment with standardized annotation semantics. An example of an annotated drawing with color-coded bounding boxes is shown in Fig. 2. The quality of drawings used for dataset construction is carefully controlled to minimize the risk of bias and drafting errors. Drawings with inconsistencies or unclear specifications are discarded. All annotations are performed by trained annotators with backgrounds in mechanical design and manufacturing and subsequently verified by two domain experts to ensure consistency and reliability of the ground truth. These measures ensure that the dataset remains both representative and trustworthy for benchmarking automation approaches, in line with recent recommendations emphasizing the risks of bias and drafting errors in engineering drawing datasets [35].

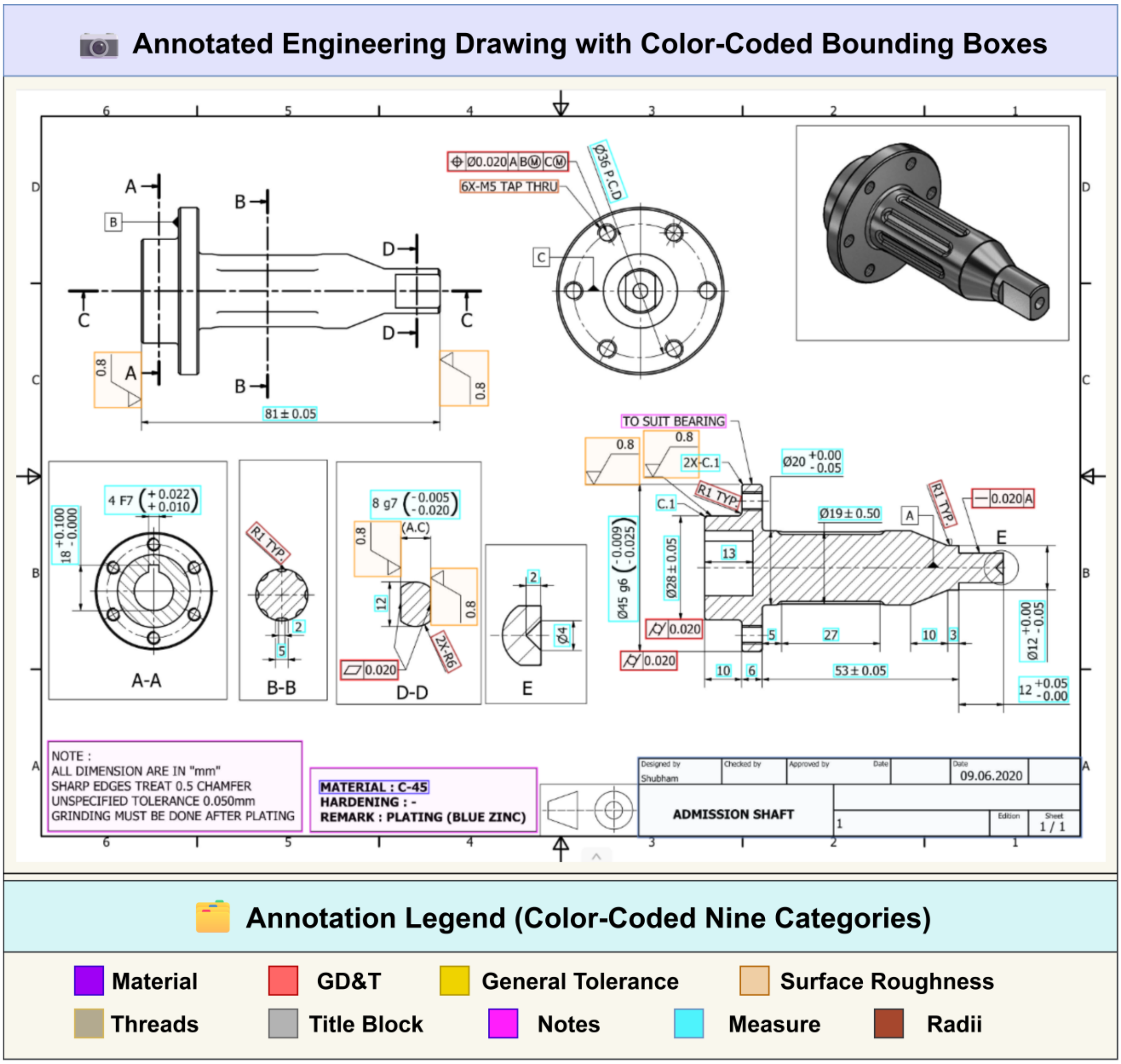

**Fig. 2.** Annotated sample from the curated dataset. Color-coded bounding boxes highlight elements across nine manufacturing-related annotation categories.

### 3.2 OBB-Based Annotation Detection and Dataset Construction

This section presents the complete pipeline for detecting annotations in 2D mechanical drawings using YOLOv11-obb, extracting image patches, and preparing a structured dataset of image-JSON pairs for downstream parsing.

#### 3.2.1 YOLOv11-obb Training

YOLOv11-obb, a one-stage object detector supporting OBBs, is employed to localize rotated and variably scaled annotations in 2D drawings. This orientation-aware detection is particularly well-suited to technical documents containing angled dimensions, skewed GD&T symbols, and vertically stacked title block entries. It is selected over other rotation-aware detectors (e.g., Oriented R-CNN [36], ReDet [37]) that rely on a two-stage design, as its one-stage architecture offers a superior trade-off between accuracy and efficiency. It achieves state-of-the-art performance on benchmarks such as DOTA-v1 (80.9 mAP) while maintaining real time inference speeds (≈10.1 ms on TensorRT) [38], making it particularly suitable for large scale industrial deployment. The model is initialized with COCO-pretrained weights [39] and fine-tuned on a training subset of the 1,367 curated engineering drawings, which collectively span diverse annotation styles, layouts, and visual conditions. The complete training configuration is provided in Table 1.

**Table 1.** YOLOv11-obb training configuration.

| Parameter | Value/Type |
|---|---|
| Model | Yolo11m-obb.pt |
| Image Size | 1024×1024 pixels |
| Epochs | 400 |
| Batch Size | 16 |
| Pretraining | COCO (Yes) |

### 3.2.2 Annotation Detection and Image Patch Extraction

Using the trained YOLOv11-obb model, annotations are detected across the full set of 1,367 engineering drawings, yielding a total of 11,469 localized annotation instances. Each detected region is classified into one of nine predefined categories described earlier in Section 4.1. Following detection, each oriented bounding box (OBB) is used to crop a rectangular image patch from the original drawing. A small contextual margin is preserved to retain relevant visual cues around the annotation. The patches are standardized in size and format and used as inputs for downstream structured parsing. The patches vary in size and content, ranging from simple dimension callouts and radii to complex GD&T frames and title block entries. On average, each drawing produces 8.4 annotation patches, depending on its complexity.

The category-wise distribution of the 11,469 extracted annotations is visualized in Fig. 3, which highlights the skewed distribution, with dominant categories including Measures, GD&Ts, and Notes, and underrepresented types such as Threads, General Tolerances, and Materials. This trend is consistent with drafting conventions: dimensions are repeated across every part or assembly view, whereas elements like general tolerances or title blocks typically occur less per drawing, and categories such as notes or surface roughness exhibit greater variability. This imbalance is addressed through targeted augmentation described in the following section.

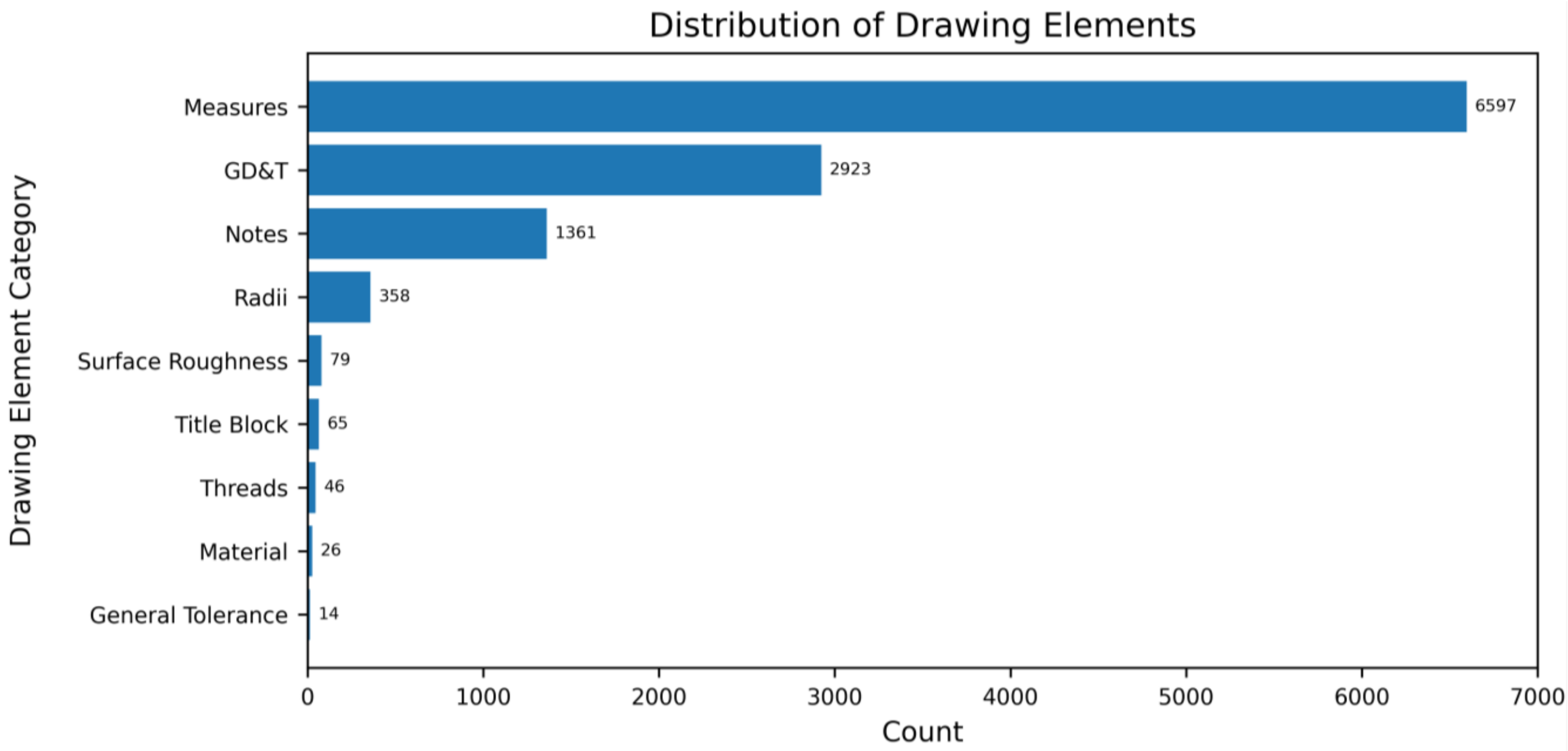

**Fig. 3.** Distribution of the 11,469 detected annotations across nine predefined categories.

To visually illustrate the detection and patch extraction process, Fig. 4 shows a sample 2D drawing with detected OBBs overlaid, along with the corresponding extracted patches organized by annotation category.

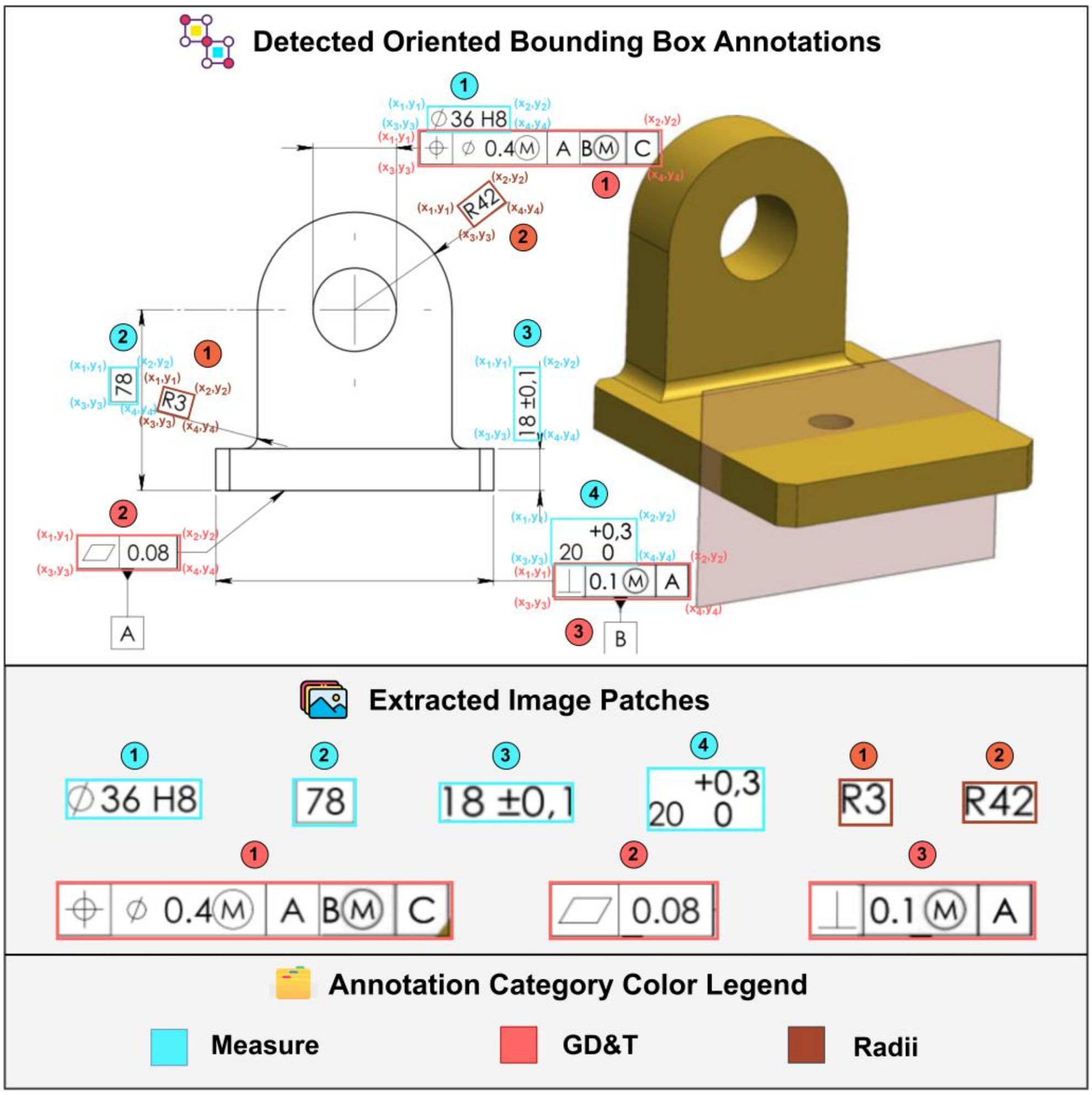

Fig. 4. Detection and patch extraction process. Sample 2D drawing with detected OBBs color-coded by annotation category (top) and corresponding extracted image patches used as inputs for structured parsing (bottom).

### 3.2.3 Annotation of OBB Patches

Each detected image patch is paired with structured JSON labels defined by category-specific schemas. These schemas are designed to represent the semantic content of the annotation in a machine-readable format suitable for training VLMs. For example:

- **GD&T** annotations include:
  1. **Geometric Characteristic**: The type of control (e.g., position, flatness)
  2. **Tolerance**: The permissible variation, often including modifiers (e.g., Ø0.020 with modifiers)
  3. **Datum Reference**: The datum features that define the tolerance frame (e.g., A, B, C)
- **Measure** annotations include:
  1. **Quantity**: The number of repeated features (e.g., 2× holes)
  2. **Nominal Value**: The intended or design-specified dimension (e.g., Ø28)
  3. **Upper/Lower Limits**: The allowable dimensional variation (e.g., ±0.05)

Fig. 5 illustrates examples of image-JSON pairs for GD&T and Measure categories, showcasing the patch and its corresponding structured label. These pairs constitute the foundational data format used for both manual supervision and downstream model training. JSON is chosen as the representation format because it is lightweight, widely adopted, and schema-constrained, allowing category-specific information to be encoded consistently while remaining interoperable with downstream reasoning engines. These pairs constitute the foundational data format used for both manual supervision and downstream model training.

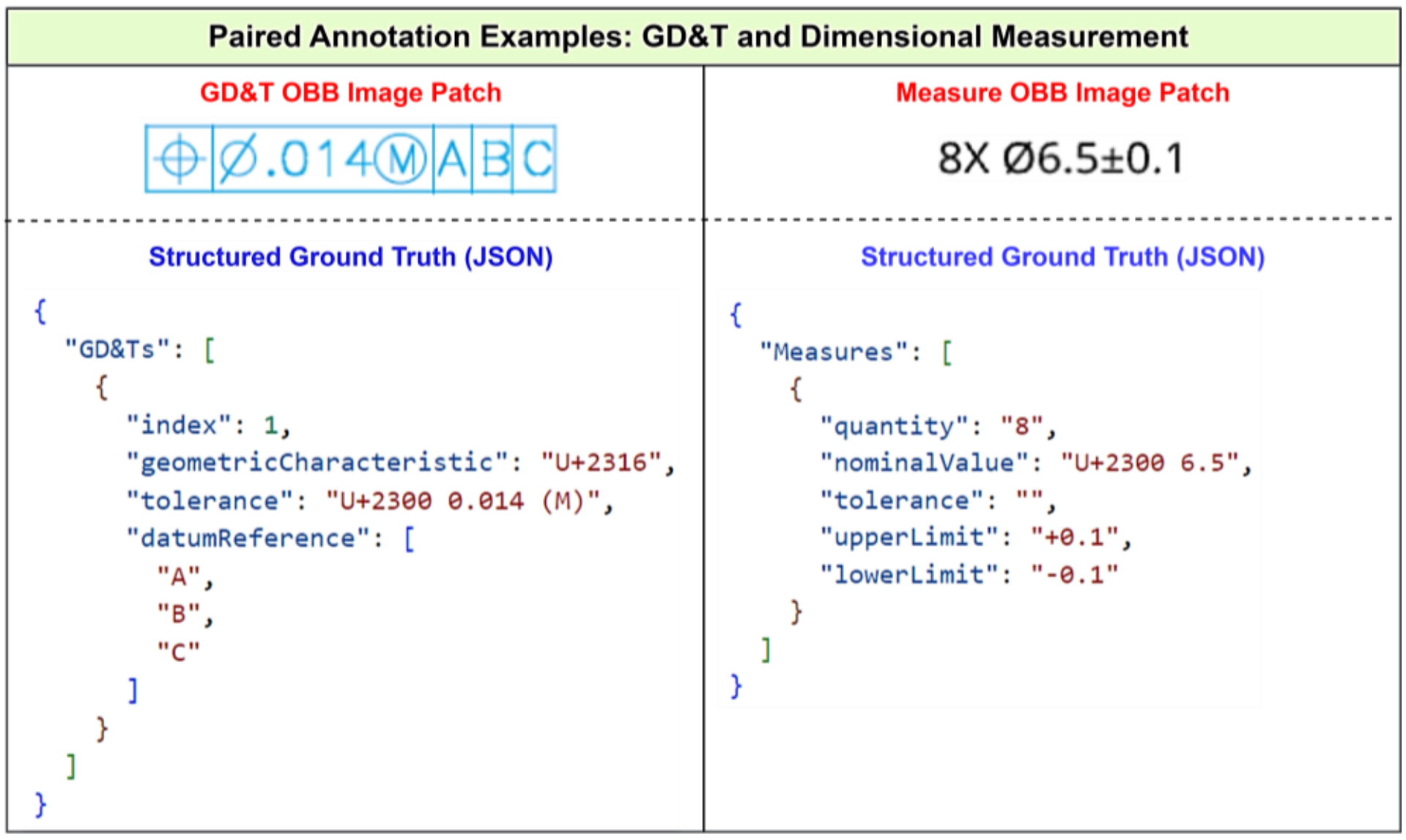

Fig. 5. Representative annotation examples for GD&T and Measure categories. Each shows an OBB-cropped image patch (top) and its structured JSON label (bottom).

To ensure consistent representation and reduce recognition errors, 14 common GD&T symbols are encoded using standardized Unicode characters in accordance with ASME Y14.5 [40]. This normalization enhances consistency, simplifies parsing, and reduces recognition errors, especially when interpreting stylized or infrequent symbols. Table 2 lists these symbols and their corresponding Unicode representations, which are integrated into the annotation schema and used consistently across the dataset, supporting a standardized approach to annotation representation and model input formatting.

**Table 2.** Unicode representations for GD&T symbols.

| Name | Symbol | Unicode |
|---|---|---|
| Position | ⌖ | U+2316 |
| Flatness | ⏥ | U+23E5 |
| Roundness | ○ | U+25CB |
| Cylindricity | ⌭ | U+232D |
| Profile of a line | ⌒ | U+2312 |
| Profile of a plane | ⌓ | U+2313 |
| Parallelism | ∥ | U+2225 |
| Perpendicularity | ⟂ | U+27C2 |
| Straightness | ⏤ | U+23E4 |
| Concentricity | ◎ | U+25CE |
| Angularity | ∠ | U+2220 |
| Symmetry | ⌯ | U+232F |
| Circular runout | ↗ | U+2197 |
| Total runout | ⌰ | U+2330 |

### 3.2.4 Data Augmentation and Final Dataset Construction

To enhance model robustness and improve generalization across varied drawing conditions, visual data augmentation is selectively applied to underrepresented annotation categories within the training set. These include General Tolerances, Material, Threads, Title Block, Surface Roughness, and Radii, categories that naturally occur less frequently compared to dominant ones like Measures and GD&Ts. The augmentation process is implemented using the PyTorch library [41], with stochastic transformation strategies designed to mimic realistic distortions in archived or scanned technical documents. The goal is to enhance diversity while maintaining semantic label consistency across image-JSON pairs. Five augmentation techniques are employed:

- **Sharpness Variation**: Simulates blur or over-sharpened scans commonly encountered in scanned technical documents
- **Contrast Adjustment**: Alters contrast levels to reflect overexposed or faded print conditions (applied with 50% probability)
- **Rotation**: Applies random 0°, 90°, 180°, or 270° orientation shifts to improve model invariance to layout orientation
- **Grayscale Conversion**: Converts colored or multi-tone drawings to monochrome (50% probability), mimicking archived blueprints
- **Color Inversion**: Inverts black and white pixels to simulate negative scans or whiteprint formats (50% probability)

A representative example is shown in Table 3, where a thread annotation image patch is augmented using all five techniques. Each transformation introduces distinct visual variation while preserving annotation structure, making the augmented data suitable for transformer-based model training.

**Table 3.** Augmentation examples applied to a thread annotation patch. Each row shows the original image (left), the augmentation type (center), and its corresponding augmented variant (right).

| Original Image Patch | Augmentation Type | Augmented Variant |
|---|---|---|
| 6X M20 X2-6H | Sharpness Variation | 6X M20 X2-6H |
| | Contrast Adjustment | 6X M20 X2-6H |
| | Rotation | 6X M20 X2-6H |
| | Grayscale Conversion | 6X M20 X2-6H |
| | Color Inversion | 6X M20 X2-6H |

These augmentations are applied exclusively to underrepresented categories to improve class balance in the training data. The effectiveness of this targeted strategy is reflected in the updated category distribution shown in Fig. 6.

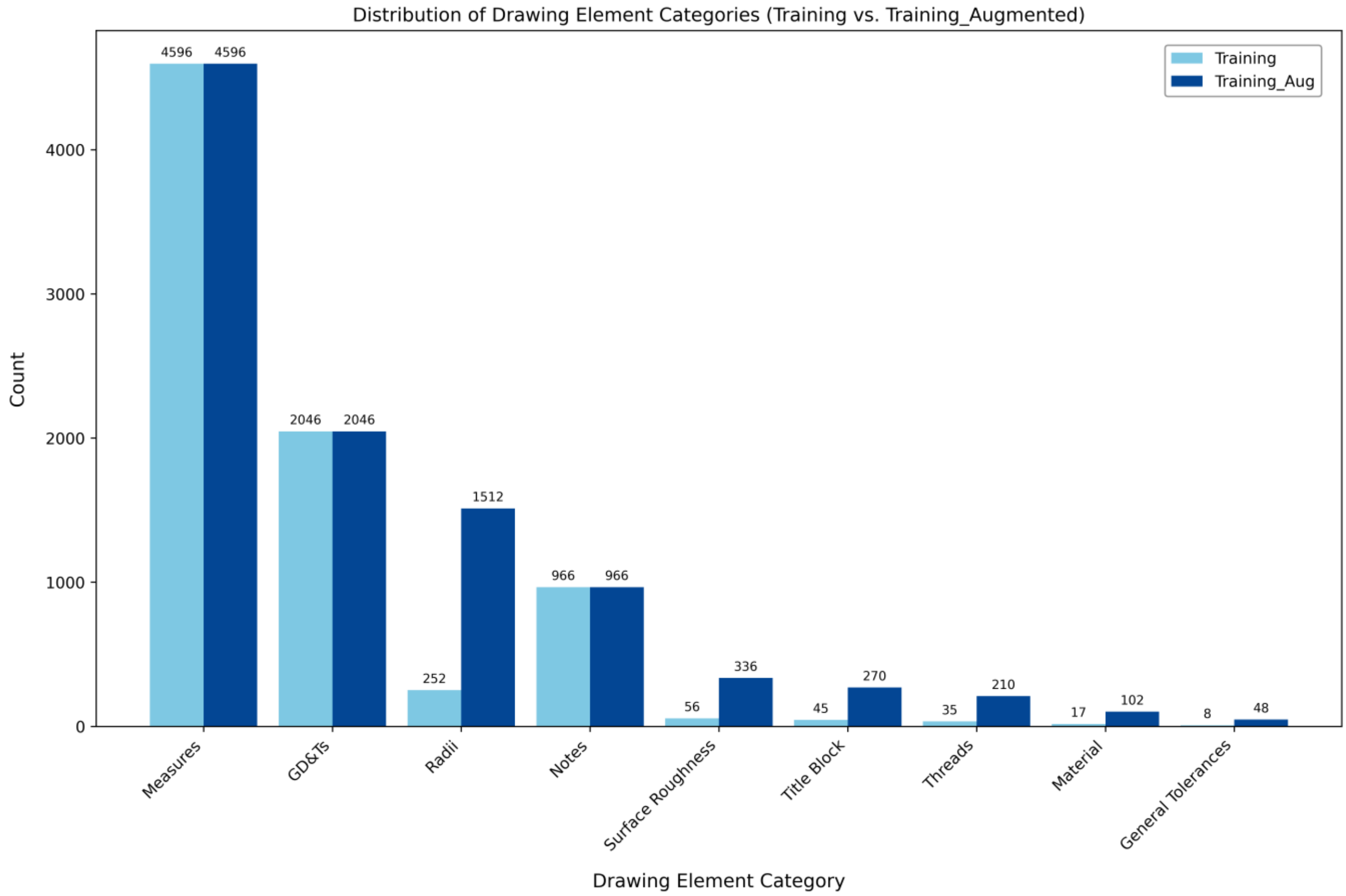

**Fig. 6.** Category-wise distribution of drawing element annotations before and after targeted augmentation. Data balance improves for minority classes such as Threads, Material, Surface Roughness, and General Tolerances.

The result is a semantically consistent, visually diverse, and class-balanced dataset of image-JSON pairs. This enriched dataset serves as the foundation for training downstream transformer-based models capable of robust, layout-invariant annotation parsing across a wide range of 2D mechanical drawings.

## 3.3 Vision-Language Models Fine-Tuning

With the finalized dataset of 11,469 image-JSON pairs, two transformer-based VLMs, Donut and Florence-2, are fine-tuned for structured parsing of engineering annotations. Both models are trained on cropped image patches containing single annotations and are trained to generate structured outputs aligned with category-specific JSON schemas. The models are selected based on their ability to jointly process visual and symbolic information, eliminate OCR dependency, and generalize across variable annotation styles, geometric layouts, and drawing artifacts common in engineering documentation.

### 3.3.1 Donut Fine-tuning

Donut is a transformer-based document parsing model with an encoder-decoder architecture capable of directly converting image inputs into structured text formats such as JSON. It operates without relying on generic OCR or region-based segmentation, making it especially well-suited for engineering drawings that contain a mix of symbolic, textual, and geometric information. In this study, Donut-base [42] is selected due to its strong performance on semi-structured documents, its ability to preserve both spatial and semantic integrity in noisy or distorted visuals, and its capacity for end-to-end fine-tuning. This OCR-free pipeline eliminates dependency on text localization and font uniformity, which are often unreliable in legacy CAD drawings, scanned blueprints, or rotated annotations. Unlike the original Donut model that supports multiple tasks such as classification, VQA, and parsing, this study exclusively focuses on parsing. Each input consists of a cropped image patch containing a single localized annotation and paired with its corresponding category-specific JSON schema. The objective is to directly generate structured outputs for downstream manufacturing without requiring additional task conditioning.

The Donut architecture follows an encoder-decoder transformer design. The visual encoder is implemented using the Swin Transformer Base (Swin-B) architecture [43], which hierarchically models spatial features via shifted window-based multi-head self-attention (MHSA). The encoder first partitions the input image patch into non-overlapping 4×4 patches and processes them through four Swin Transformer stages, configured with {2, 2, 14, 2} transformer layers and a window size of 10, as originally defined in Donut. Each Swin block consists of layer normalization, GELU activations, two-layer MLPs, and window-based MHSA using key-query-value (KQV) attention. The encoder converts the input image patch (cropped engineering annotation patch) into a sequence of 1024-dimensional latent visual embeddings, which serve as conditioning inputs to the decoder.

The textual decoder is initialized from the pre-trained multilingual BART model [44], consistent with the original Donut implementation. The decoder autoregressively generates output tokens corresponding to structured JSON fields aligned with the annotation schema. Prompt tokens are applied during inference to condition decoding for category-specific parsing. The decoder employs masked multi-head self-attention, encoder-decoder cross-attention, and standard feed-forward layers with GELU activations. During fine-tuning, the full encoder-decoder model is jointly trained end-to-end using cross-entropy loss over the output token sequences. In total, approximately 143 million parameters are optimized, consistent with the original Donut configuration. No architectural modifications are introduced; only the task objective is adapted to parsing-only mode for structured engineering annotation extraction. The complete parsing pipeline is shown in Fig. 7.

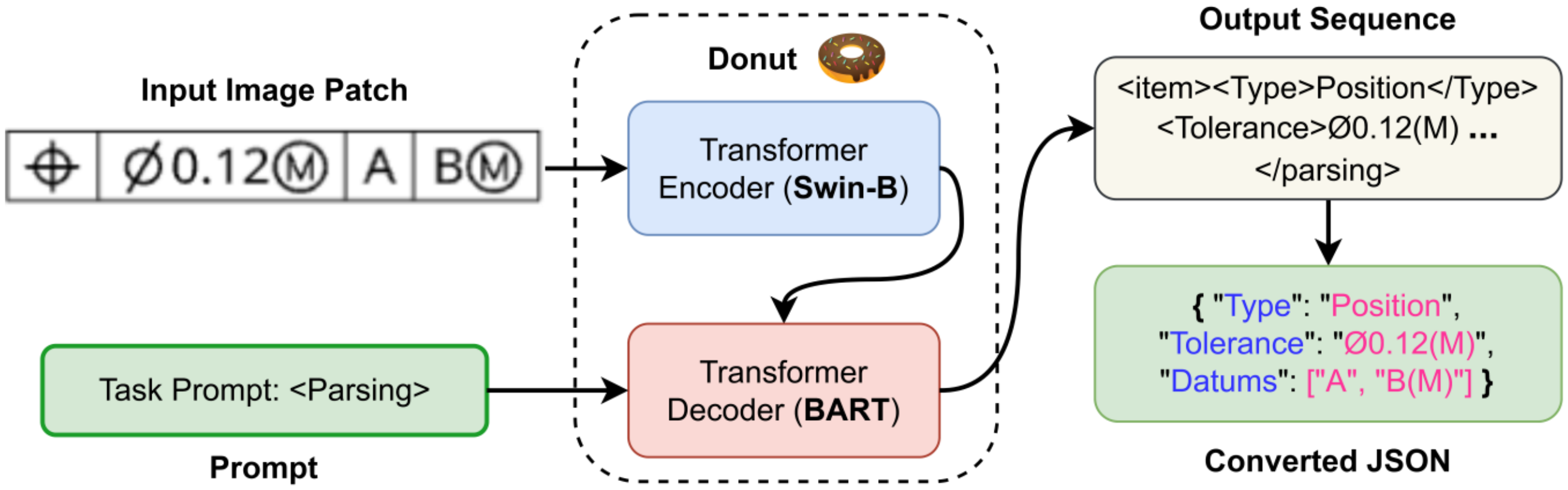

**Fig. 7.** Donut architecture tailored for engineering annotation parsing. The Swin Transformer encoder extracts hierarchical visual

embeddings from input image patches containing individual annotations. The BART-based decoder generates structured JSON outputs under prompt conditioning, producing machine-readable annotations for downstream manufacturing tasks [10].

Two modeling strategies are explored during fine-tuning:

- **Unified model:** A single Donut model is trained across all nine annotation categories using a shared structured output format. Each image patch is paired with its category-specific JSON label, and the model learns to handle diverse annotation types and layouts within one unified architecture. This approach promotes generalization across classes and simplifies deployment by consolidating all inference into a single model.
- **Category-specific models:** Nine independent Donut models are fine-tuned separately, each restricted to one annotation category and its corresponding schema. This strategy emphasizes tailored learning for each annotation type but requires training, managing, and deploying multiple models, one per category.

The differences between these two strategies are illustrated in Fig. 8. In the unified setup (Fig. 8a), a single Donut model ingests a combination of image and JSON ground truth, regardless of annotation type, and outputs structured data using a general schema. In contrast, the category-specific strategy (Fig. 8b) assigns a distinct model to each annotation type (e.g., GD&T, Measure, Thread), each trained only on its respective subset. To maintain clarity, only three annotation categories are illustrated in Fig. 8, while the remaining six, following the same structure, are omitted for simplicity. Based on our prior findings in [45], the unified model consistently outperforms category-specific models in terms of generalization, training efficiency, and deployment simplicity. It further reduces redundancy by enabling shared learning of visual and structural patterns across categories. Therefore, only the unified Donut model is adopted in this study for final evaluation. The complete fine-tuning configuration used for both Donut models is summarized in Table 4.

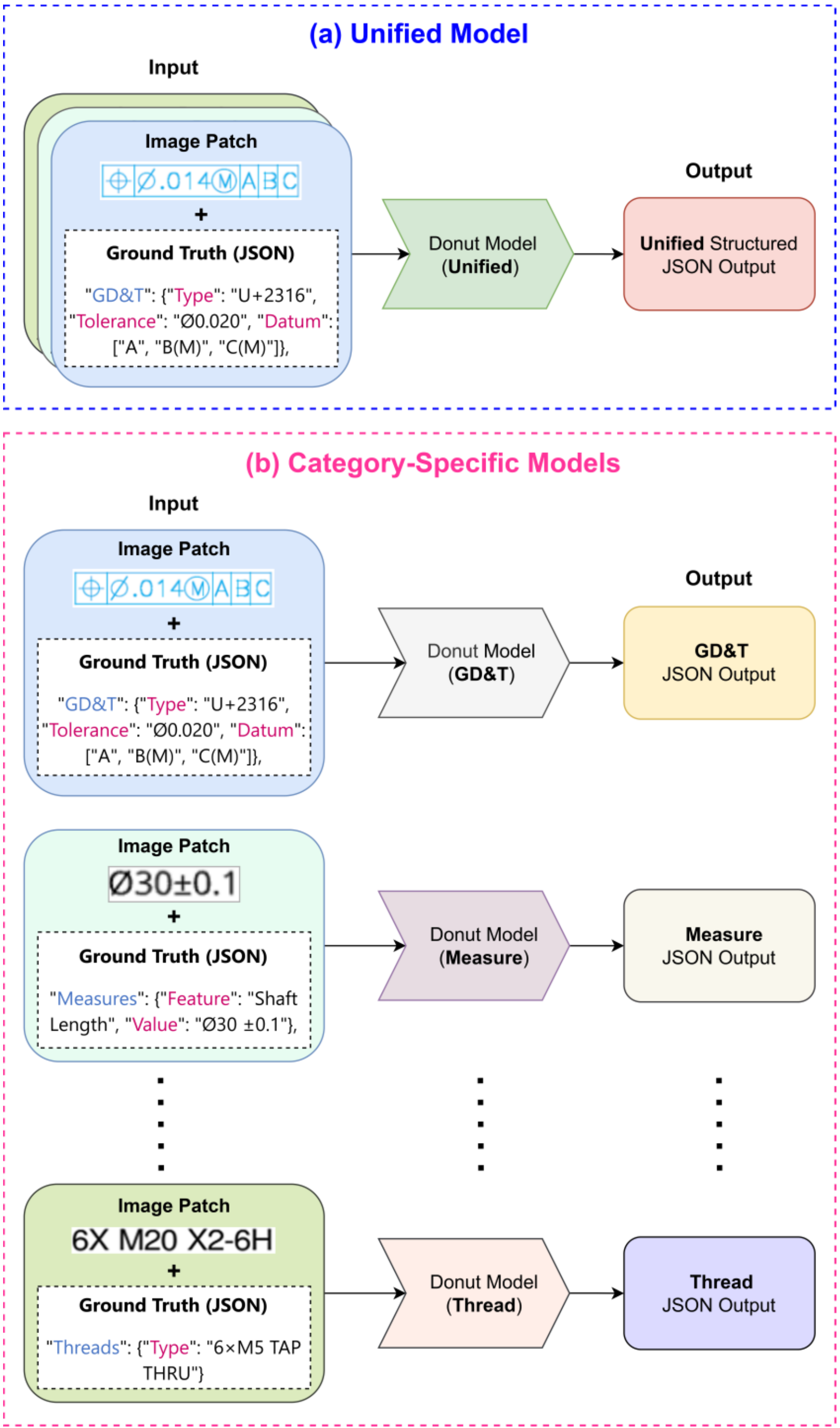

**Fig. 8.** Donut fine-tuning strategies: unified vs. category-specific. (a) In the unified setup, a single Donut model is trained on all annotation categories using a shared structured JSON output. (b) In the category-specific approach, individual models are fine-tuned per annotation type. Three representative categories (GD&T, Measure, Thread) are shown; the rest are indicated with dotted extensions.

### 3.3.2 Florence-2 Fine-tuning

Florence-2 is a transformer-based VLM with an encoder-decoder architecture designed to unify diverse visual understanding tasks into a single prompt-driven generative framework. The architecture processes both visual and textual inputs to generate structured text outputs directly, without relying on OCR modules or region-specific detectors. This OCR-free and end-to-end generative design makes Florence-2 especially well-suited for structured parsing of engineering drawings, where annotation styles, geometric layouts, and visual artifacts frequently vary across scanned blueprints, CAD exports, and legacy documentation.

In this study, Florence-2-base (0.23 billion parameters) [46] is selected for fine-tuning due to its efficient model size and strong generalization capability across visually complex structured domains. The architecture consists of two main modules: a vision encoder and a multi-modal transformer encoder-decoder. The vision encoder is implemented using the DaViT (Dual Attention Vision Transformer) backbone [47], which hierarchically encodes spatial features through both spatial and channel-wise attention mechanisms. The encoder partitions each cropped annotation image patch into non-overlapping patches, which are then processed through DaViT stages with embedding dimensions of [128, 256, 512, 1024], transformer block configurations of [1, 1, 9, 1], and attention heads [4, 8, 16, 32]. This hierarchical design converts the input patch into a sequence of latent visual embeddings.

The multi-modal transformer encoder-decoder builds upon a pre-trained BART model, wherein the text embeddings and vision embeddings are unified into a shared representation space. The encoder consists of 6 transformer layers with 768-dimensional embeddings, while the decoder similarly comprises 6 transformer layers of identical embedding size. The transformer layers include masked multi-head self-attention, encoder-decoder cross-attention, and standard feed-forward layers. Location embeddings are incorporated into the token sequence by quantizing spatial coordinates into 1,000 discrete bins, following the approach described in Pix2Seq-like formulations [48]. This enables the model to represent bounding boxes, quadrilaterals, and polygonal regions uniformly as sequences of location tokens for spatially grounded parsing tasks. The full Florence-2-base model contains approximately 232 million parameters and is fine-tuned end-to-end with cross-entropy loss on output token sequences aligned with category-specific structured JSON schemas. The complete model architecture employed for engineering annotation parsing is illustrated in Fig. 9.

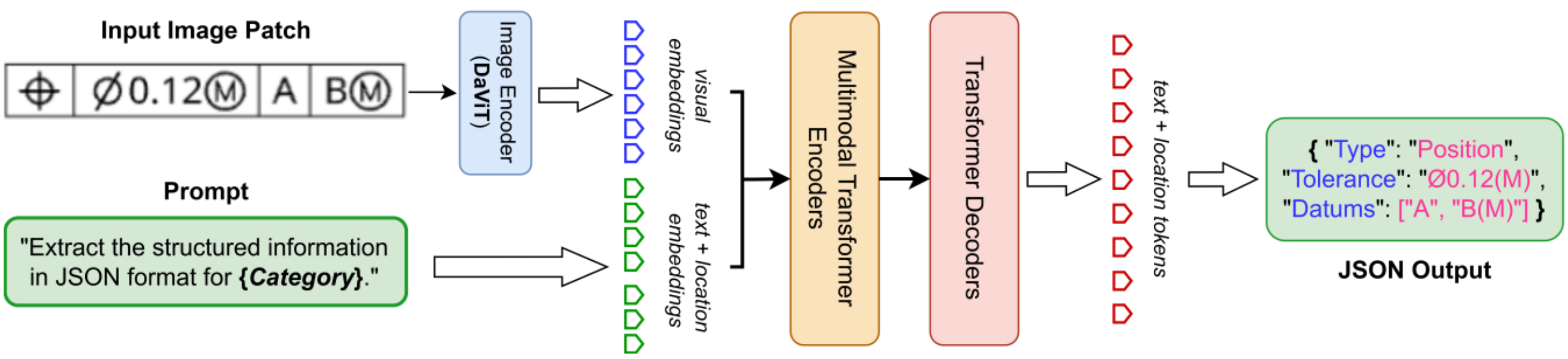

**Fig. 9.** Florence-2 architecture for structured engineering annotation parsing. The DaViT encoder extracts hierarchical visual embeddings from cropped annotation patches, which are fused with prompt embeddings through a multimodal transformer encoder-decoder to generate structured JSON outputs [11].

Fine-tuning is formulated as a prompt-based structured generation task, where each training instance consists of a cropped image patch containing a single annotation and an accompanying natural language prompt. The prompt follows the template: *Extract the structured information in JSON format for {Category}*, where

{Category} is dynamically replaced with one of the nine annotation types. This formulation enables Florence-2 to generate outputs conditioned on both visual content and prompt intent, facilitating structured information extraction aligned with the correct annotation schema. The full architecture of this fine-tuning pipeline is illustrated in Fig. 10, where each training pair (image patch and JSON label) is processed end-to-end by the model. Florence-2 undergoes full parameter fine-tuning [49], where all model weights are updated during training to maximize consistency between predicted and reference structured outputs and to effectively learn task-specific visual and semantic patterns.

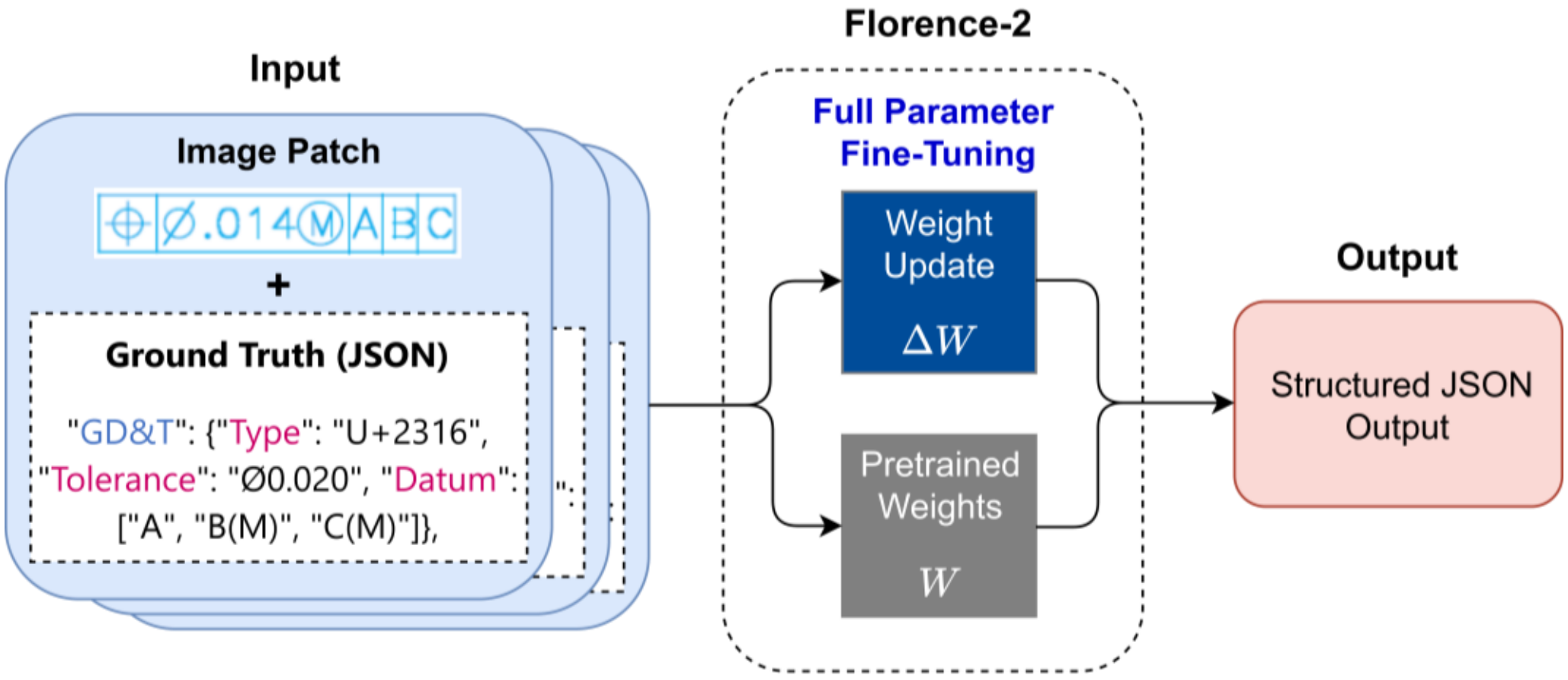

**Fig. 10.** Florence-2 fine-tuning pipeline for structured annotation parsing.

To ensure a fair comparison with Donut, Florence-2 is trained using an identical configuration. Each instance includes an image patch paired with its structured label. The shared configuration is summarized in Table 4.

**Table 4.** Shared fine-tuning configuration used for both Donut and Florence-2 models.

| Parameter | Value/Type |
|---|---|
| Models | Donut-base & Florence-2-base |
| Optimizer | AdamW (Cosine decay) |
| Learning Rate | 1e-6 |
| Batch Size | 1 |
| Epochs | 30 |
| Loss Function | Cross-entropy |

## 4. Results and Discussion

This section presents the quantitative and qualitative evaluation of the proposed hybrid vision-based parsing framework, including (1) object detection performance of the YOLOv11-obb model, (2) structured parsing results of the fine-tuned Donut and Florence-2 models, and (3) real-world validation via GUI-based semantic overlays. The overall results demonstrate strong performance in both detection and structured understanding tasks, validating the proposed system's applicability in knowledge-driven manufacturing workflows.

### 4.1 YOLOv11-obb Detection Performance

During inference, the trained YOLOv11-obb model demonstrates strong capability in detecting diverse annotation types across 2D mechanical drawings. As shown in Fig. 11, the evaluation metrics converge with high stability, with final precision, recall, mAP@0.5, and mAP@0.5–0.95 all consistently exceeding 0.95. The dataset is split 90:10 at the drawing level, ensuring that test samples came from previously unseen drawings across diverse formats. Evaluation on this held-out set confirms robust generalization, with all annotation categories achieving confidence scores above 95%. This highlights the model's reliability even in cluttered or rotated annotation settings, a critical requirement for engineering drawings where annotation angles and styles vary significantly.

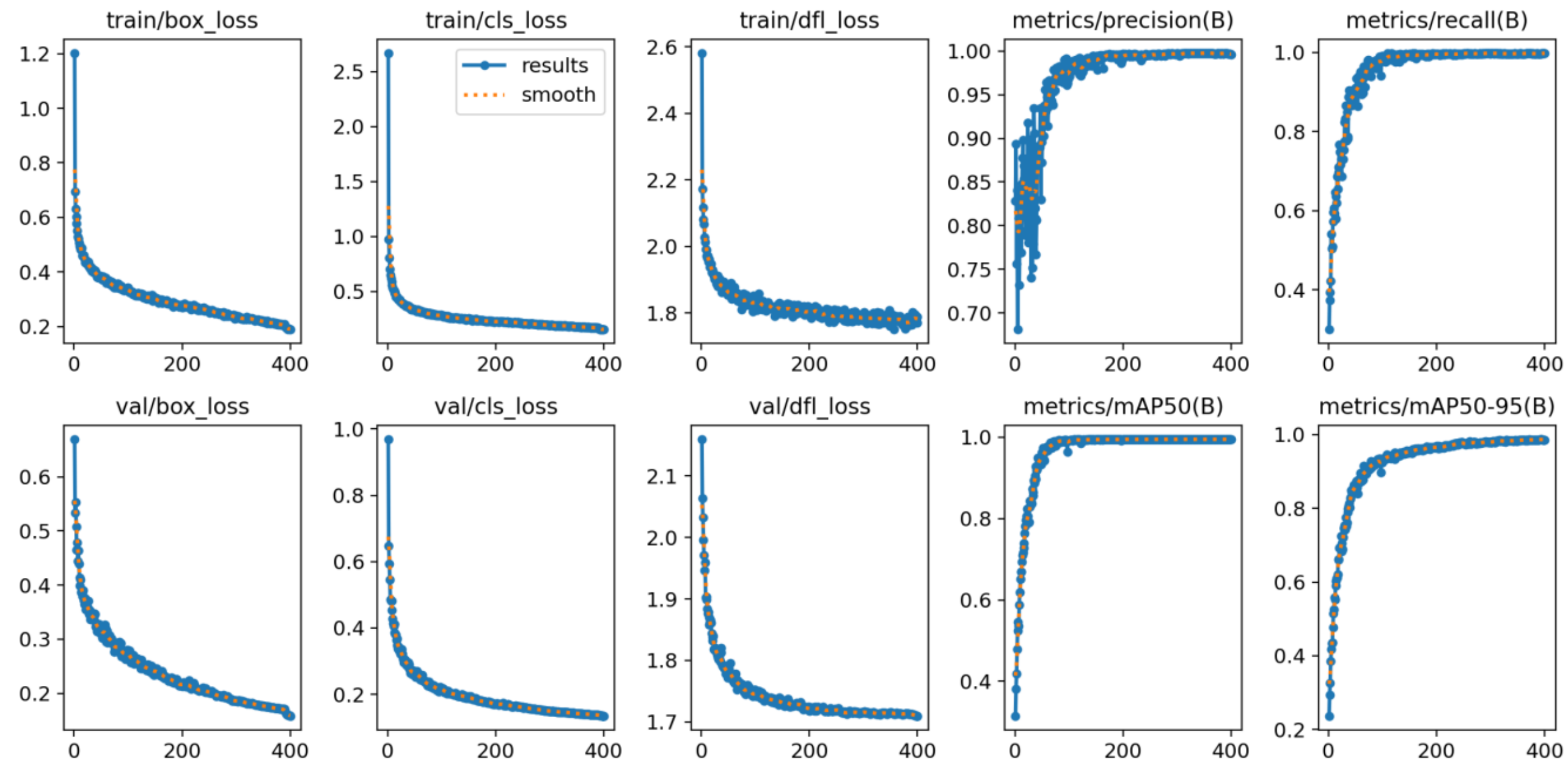

**Fig. 11.** Performance curves for YOLOv11-obb across key metrics: bounding box loss, classification loss, DFL loss, precision, recall, mAP@0.5, and mAP@0.5–0.95.

A representative detection outcome is visualized in Fig. 12, where the original input drawing is overlaid with predicted annotation regions, categorized and color-coded with associated confidence scores. The output demonstrates robust performance even in dense annotation environments, accurately segmenting overlapping elements such as surface roughness symbols, GD&T frames, dimension callouts, and title block entries.

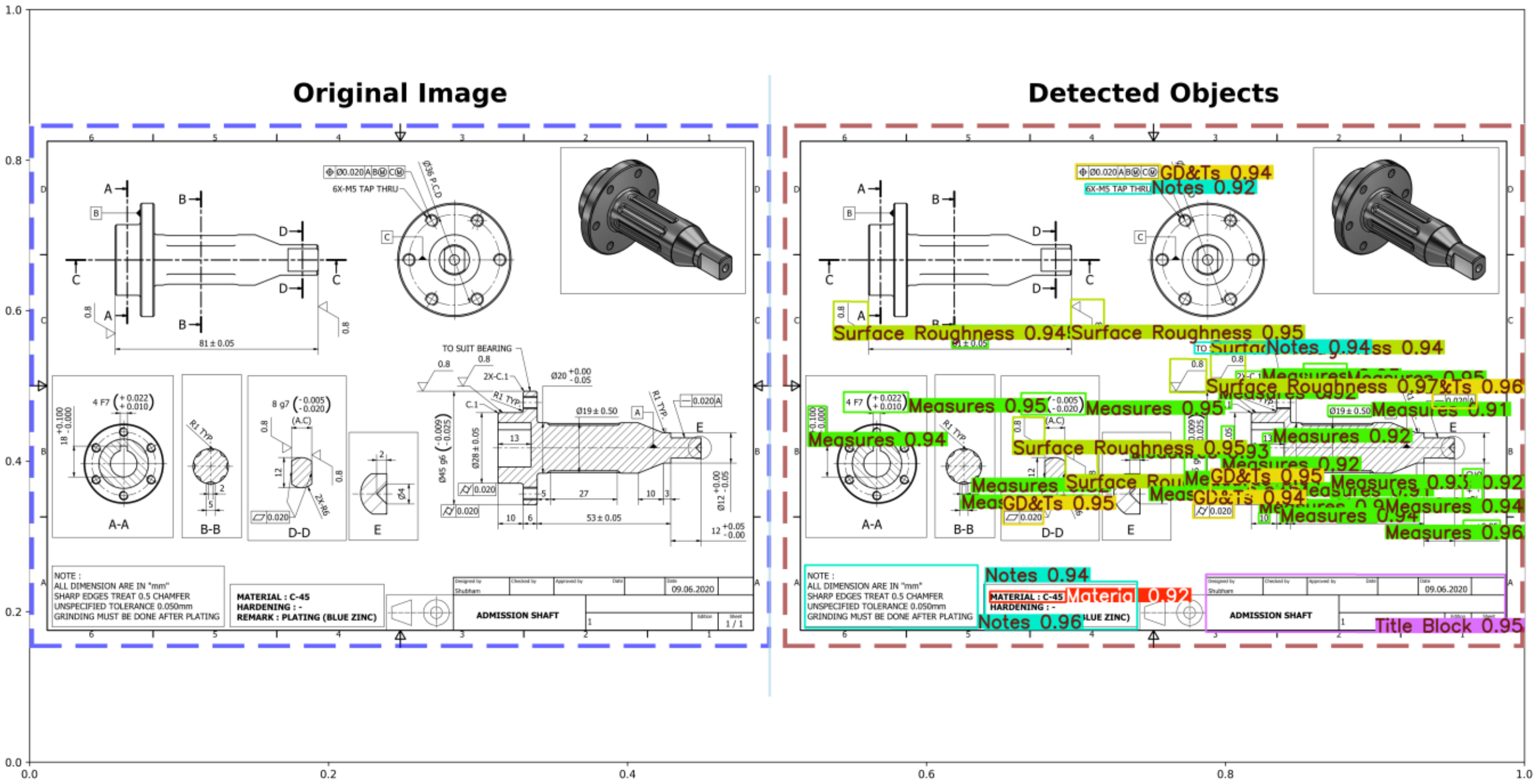

**Fig. 12.** Sample detection result from YOLOv11-obb. Left: original engineering drawing. Right: detected annotation regions overlaid with category labels and confidence scores.

A more granular evaluation is presented through confusion matrices in Fig. 13(a) (normalized) and Fig. 13(b) (raw counts). The normalized matrix highlights near-perfect classification accuracy for almost all categories with scores approaching 1.0.

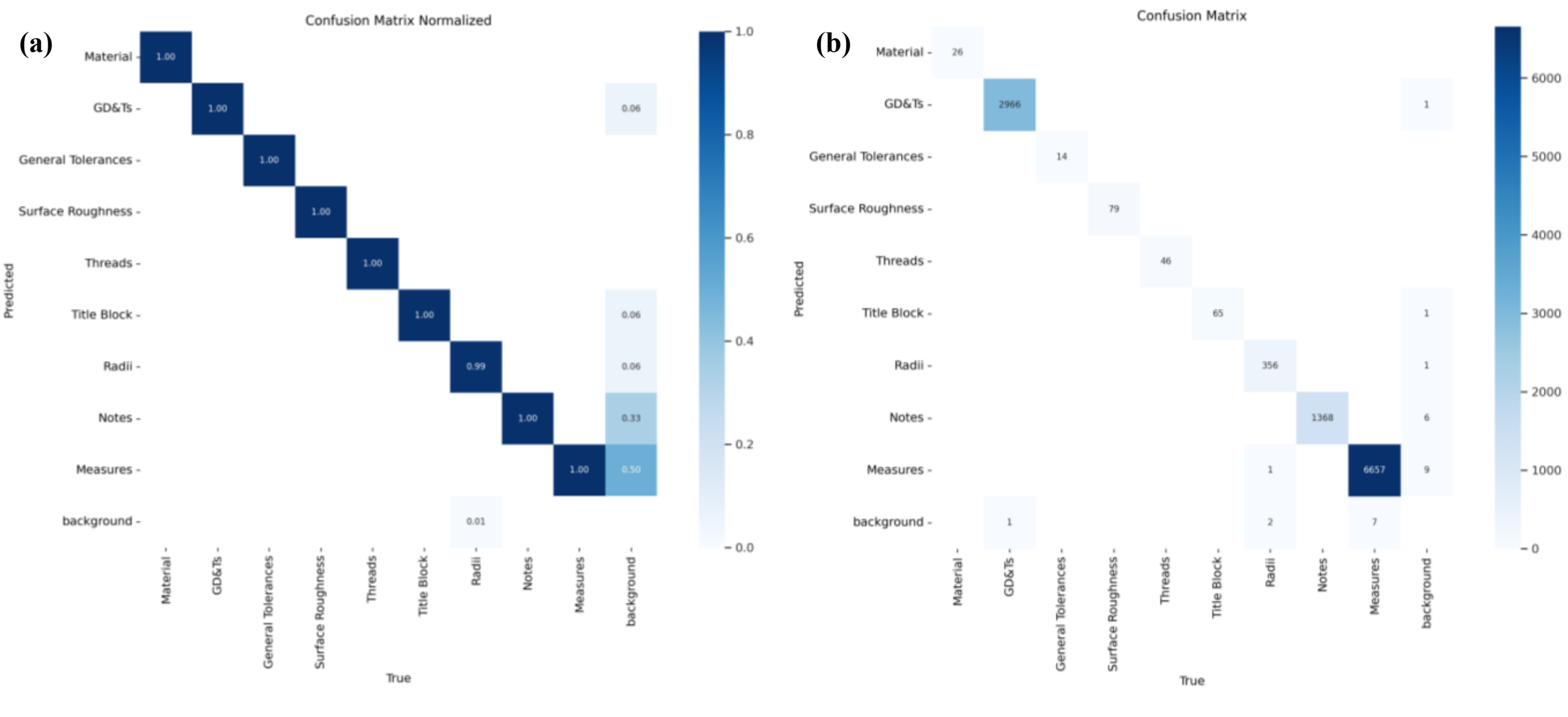

**Fig. 13.** Confusion matrix evaluation for YOLOv11-obb. (a) Normalized accuracy matrix showing high class separability. (b) Raw prediction counts highlighting class frequency and minor misclassifications.

The raw confusion matrix additionally confirms the long-tail class distribution within the dataset, with categories such as Material, Threads, and General Tolerances occurring less frequently. This imbalance influences class frequency but has minimal impact on detection accuracy, as precision and recall remain high even for these sparse classes. The model's consistent classification fidelity, despite this imbalance, demonstrates its robustness under real-world annotation diversity. These results collectively confirm the suitability of YOLOv11-obb as the detection backbone in the proposed framework. It achieves strong localization and classification performance under varying drawing layouts, providing a reliable basis for subsequent structured parsing and downstream applications.

### 4.2 Structured Parsing Performance

Following the annotation localization performed by YOLOv11-obb, each detected region is semantically parsed using one of two fine-tuned VLMs: Donut or Florence-2. The outputs are structured JSON fields specific to each annotation category. To evaluate model performance at inference time, the predicted JSON outputs are compared against manually verified ground truth on a test subset comprising 10% of the full dataset, ensuring representative coverage across all nine annotation types.

Evaluation is conducted at the OBB patch level, with each prediction assessed independently through field-level comparisons. For each annotation patch, the number of True Positives (TP), False Positives (FP), and False Negatives (FN) is computed based on exact key-value matches between the predicted and ground truth JSON fields. The definitions are as follows:

- **TP:** A predicted key-value pair that exactly matches the corresponding ground truth.
- **FP:** A predicted key-value pair that is either incorrect or not present in the ground truth.
- **FN:** A ground truth key-value pair that is missing from the prediction.

Based on these counts, four evaluation metrics are computed for each annotation category and then aggregated to produce overall scores. These metrics are calculated per annotation class and subsequently aggregated to report both class-wise and overall performance. The metrics used are:

$$\text{Precision} = \frac{\text{TP}}{\text{TP} + \text{FP}} \tag{1}$$

$$\text{Recall} = \frac{\text{TP}}{\text{TP} + \text{FN}} \tag{2}$$

$$\text{F1 score} = 2 \times \frac{\text{Precison } \times \text{Recall}}{\text{Precison} + \text{Recall}} \tag{3}$$

$$\text{Hallucination Rate} = 1 - \text{Precision} \tag{4}$$

The hallucination rate is particularly important in engineering contexts, where over-generation of semantically invalid or incorrect fields can cause misinterpretation, downstream errors, or regulatory non-compliance. A hallucinated field is defined as any prediction not aligned with a valid key-value pair in the ground truth, reflecting poor semantic discipline in model output.

Table 5 summarizes the parsing results for Donut and Florence-2 across all annotation categories on the test data. Donut consistently demonstrates higher accuracy, achieving an overall F1-score of 94%, precision of 89.2%, recall of 99.2%, and a low hallucination rate of 10.8%. In contrast, Florence-2 achieves an F1-score of 85.0%, with 78.4% precision, 92.7% recall, and a higher hallucination rate of 21.6%. These results highlight the comparative advantage of Donut's encoder–decoder architecture in enforcing schema conformity and mitigating overgeneration.

**Table 5.** Structured parsing performance on the test set (10% of 11,469 image patches) across nine annotation categories.

| Category | Donut | | | | Florence-2 | | | |
|---|---|---|---|---|---|---|---|---|
| | Precision | Recall | F1 score | Hallucination | Precision | Recall | F1 score | Hallucination |
| **Measures** | 0.896 | 0.992 | 0.941 | 0.104 | 0.76 | 0.873 | 0.813 | 0.24 |
| **Title Block** | 0.522 | 0.545 | 0.533 | 0.478 | 0.302 | 0.52 | 0.382 | 0.698 |
| **GD&Ts** | 0.933 | 1.0 | 0.965 | 0.067 | 0.838 | 0.995 | 0.91 | 0.162 |
| **Notes** | 0.681 | 1.0 | 0.81 | 0.319 | 0.655 | 1.0 | 0.791 | 0.345 |
| **Material** | 1.0 | 1.0 | 1.0 | 0.0 | 1.0 | 1.0 | 1.0 | 0.0 |
| **Surface Roughness** | 1.0 | 1.0 | 1.0 | 0.0 | 0.857 | 0.923 | 0.889 | 0.143 |
| **Radii** | 0.891 | 1.0 | 0.943 | 0.109 | 0.837 | 0.818 | 0.828 | 0.163 |
| **Threads** | 0.833 | 0.909 | 0.870 | 0.167 | 0.75 | 0.6 | 0.667 | 0.25 |
| **General Tolerance** | 0.5 | 1.0 | 0.667 | 0.5 | 0.5 | 1.0 | 0.667 | 0.5 |
| **Overall** | **0.892** | **0.992** | **0.940** | **0.108** | **0.784** | **0.927** | **0.85** | **0.216** |

A closer look at category-wise results reveals distinct behavioral patterns. Frequent and structurally consistent categories such as Measures and GD&Ts exhibit high recall and F1-scores for both models. However, Donut shows higher schema compliance and token-level accuracy, particularly in symbol-heavy formats such as GD&T frames, where Unicode-encoded control characteristics and datum references must be parsed precisely. Symbolically constrained categories such as Surface Roughness and Radii, although less frequent, are parsed with high fidelity, achieving perfect or near-perfect F1-scores with Donut. These categories benefit from visual regularity and consistent layout cues, allowing the models to generalize despite limited exposure during fine-tuning.

In contrast, categories with lower visual structure or inconsistent formatting, such as Title Block, General Tolerances, and Notes, pose greater challenges. Both models demonstrate perfect recall on Notes, but their lower precision leads to high hallucination rates, indicating a tendency to over generate in the presence of free-form or overlapping content. Title Block performance is further hindered by its tabular density and semantic overlap with adjacent elements like material specifications and notes, causing frequent misalignments between predicted fields and the ground truth schema. Florence-2, in particular, shows higher volatility in these ambiguous regions, likely due to its broader pretraining and lack of enforced structural decoding.

The consistent edge shown by Donut can be attributed to its decoder-centric architecture and task-specific fine-tuning that emphasize structured output alignment. Its OCR-free pipeline allows it to handle rotated or distorted text with minimal dependency on explicit token localization. Florence-2, while lightweight and adaptable through prompt-based decoding, suffers from reduced control over output format in categories that lack visual regularity. Nonetheless, both models successfully demonstrate the feasibility of transformer-based parsing in a multimodal setting, converting unstructured drawing annotations into actionable semantic representations.

Overall, these results validate the design of a modular two-stage pipeline, where high-precision object detection is followed by robust semantic parsing. The performance trends observed across annotation categories offer valuable insights for future improvements. Categories with consistent symbolic layouts benefit from end-to-end visual learning, whereas free-form and layout-dense regions may require hybrid approaches that combine transformer outputs with schema-constrained post-processing or rule-based verification to ensure reliability in production contexts.

### 4.3 Qualitative Validation and Practical Readiness

While the preceding quantitative results validate the effectiveness of the proposed framework in structured parsing of engineering annotations, qualitative analysis further illustrates its utility in real-world scenarios. To support this, a custom-built graphical user interface (GUI) is developed to visualize the full inference pipeline, from annotation detection to structured semantic parsing, on 2D mechanical drawings. The GUI allows users to upload a 2D drawing and view annotated overlays in real time, along with category-wise confidence scores and extracted semantic fields. As shown in Fig. 14(a), a representative input drawing contains diverse annotation types including dimensional callouts, GD&T feature frames, surface roughness symbols, material specifications, and free-text notes. The model predictions are overlaid in Fig. 14(b), with each detected region color-coded by annotation category and labeled with its confidence score. This overlay provides an intuitive visualization of how the model segments the drawing into meaningful regions and associates semantic labels with visual elements.

**(a)**

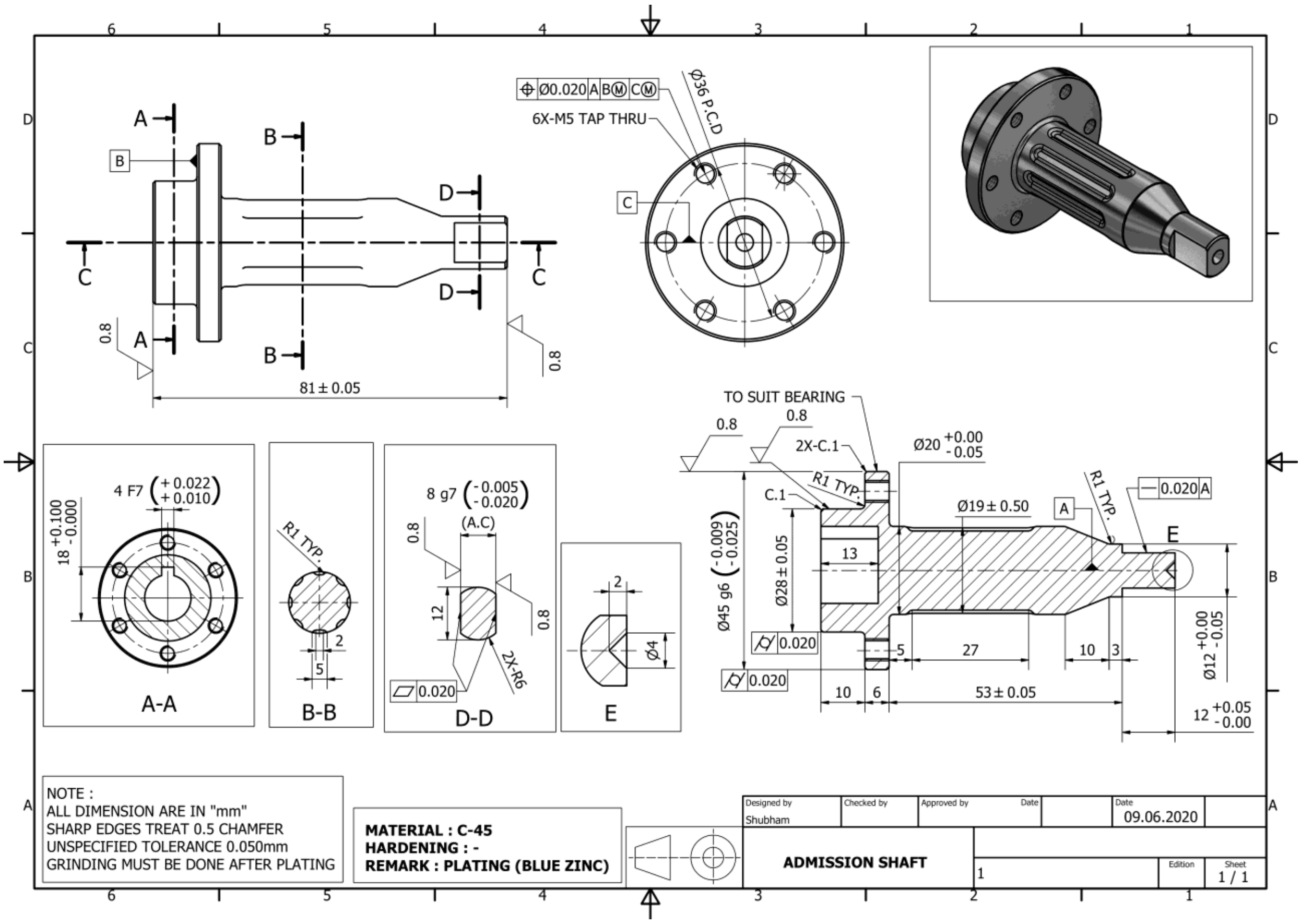

**(b)**

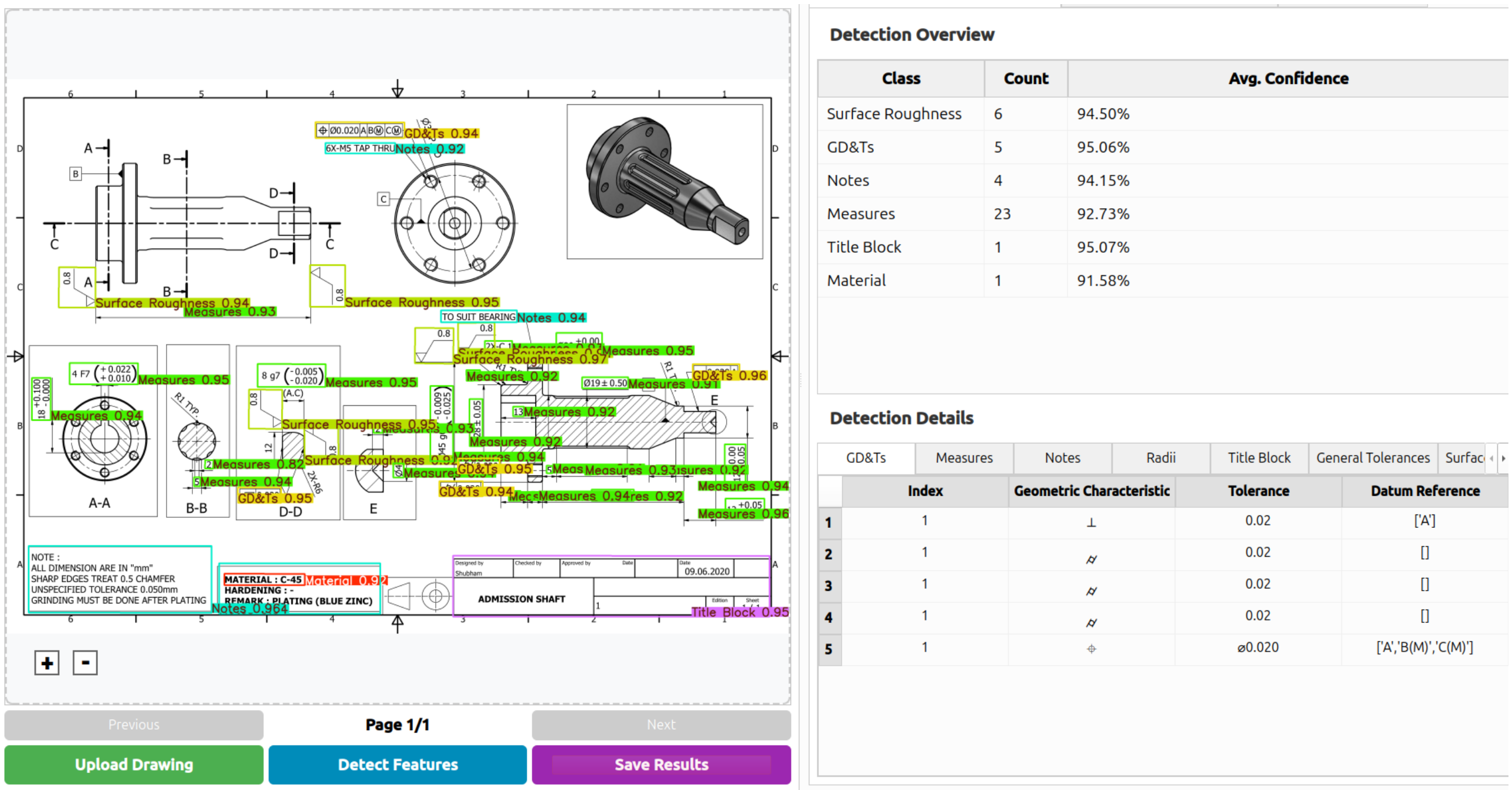

**Fig. 14.** Graphical interface for qualitative parsing validation. (a) Original engineering drawing. (b) Parsed output showing detected

annotations, predicted categories, and confidence scores.

Beyond visual overlays, the GUI includes a structured tabular interface that displays the parsed fields organized by category. For example, in the Measures section, features such as shaft length, diameter, and slot size are listed with their corresponding values and tolerances. Similarly, GD&T annotations are decoded into structured fields including geometric characteristics, tolerance zones, and referenced datums. This interactive presentation bridges visual localization with semantic understanding, enabling users to audit, interpret, and export results for downstream workflows.

To assess output consistency and integration readiness, the system also provides exportable structured JSON files containing the full parsed content. As shown in Fig. 15, this output captures all annotation categories in a machine-readable format, suitable for integration with CAD/CAM environments, rule-based process planning systems, or inspection report generation pipelines. Key annotation fields are presented with standardized keys and cleanly extracted values, aligned with the category-specific schemas described earlier in the methodology. Additional GUI-based parsing results on diverse engineering drawings are included in **Appendix A** to further demonstrate the system's generalization across varying annotation layouts and drawing complexities.

```
{
"Material": "C-45",
"Threads": [{"Type": "6×M5 TAP THRU"}],
"GD&T": [
   {"Type": "Position", "Tolerance": "Ø0.020", "Datums": ["A", "B(M)", "C(M)"]},
   {"Type": "Straightness", "Tolerance": "0.020", "Datums": ["A"]},
   {"Type": "Cylindricity", "Tolerance": "0.020", "Datums": []},
   {"Type": "Flatness", "Tolerance": "0.020", "Datums": []}],
"General Tolerance": "",
"Radii": "",
"Surface Roughness": [
   {"Ra": "0.8 µm"}],
"Measures": [
   {"Feature": "Shaft Length", "Value": "81 ±0.05 mm"},
   {"Feature": "Diameter", "Value": "Ø28 ±0.05 mm"},
   {"Feature": "Slot", "Value": "2×4 mm"}],
 "Title Block": {
   "Designer": "Shubham",
   "Date": "09.06.2020",
   "Drawing Name": "Admission Shaft"},
"Notes": "All dimensions are in mm.
      Sharp edges treat 0.5 chamfer.
      Unspecified Tolerance 0.050 mm.
      Grinding must be done after plating."
}
```

**Fig. 15.** Example of structured JSON output covering all semantic fields extracted from the drawing.

Qualitatively, the system demonstrates strong generalization across annotation densities, drawing styles, and visual conditions. The results also reveals subtle performance nuances, such as slight over-segmentation in densely packed title blocks or minor field hallucinations in highly variable categories. These observations reinforce the distinct operational behaviors of the underlying models. Donut, with its encoder–decoder structure, generates more schema-compliant and structured outputs, which is an essential feature for downstream applications such as manufacturing automation. Florence-2, while more flexible and lightweight, exhibits a higher tendency toward overgeneration in layout-dense or free-text categories like Title Block and Notes. While such outputs may carry semantically relevant content, they sometimes deviate from strict annotation schemas required for engineering validation. These limitations, especially in symbol-heavy or underrepresented categories, underscore the need for post-processing validation and schema-constrained inference during production deployment. Improvements may come from integrating lightweight rule-based verification, category-aware decoding heads, or hybrid retrieval-augmented workflows to bolster semantic discipline and reduce hallucination. It should also be noted that strict semantic interpretation of annotations requires linking annotations to reference geometry and views. The current framework focuses on detection and structured representation of annotation content, while view-annotation integration and deeper semantic reasoning are recognized as important directions for future work.

Overall, the GUI-based validation confirms that the proposed pipeline operates reliably not only in controlled evaluations but also in complex, symbol-rich, industrial-grade drawings. The combination of high-confidence detection, structured semantic decoding, and user-facing interpretability highlights the framework's readiness for deployment in real-world design, inspection, and manufacturing environments. Together, these findings validate the broader use of fine-tuned transformer-based models for structured information extraction in engineering, while also charting a clear path for enhancing robustness in low-data or semi-structured settings. By transforming previously unstructured drawing content into actionable digital representations, the system supports knowledge-driven engineering processes and paves the way for tighter integration into digital thread ecosystems.

## 5. Case Study: Rule-Based Interpretation of Extracted Drawing Information for Digital Manufacturing

The structured annotation data extracted by the proposed pipeline is intended not only for accurate interpretation but also for seamless integration into downstream manufacturing workflows. As illustrated in Fig. 16, these structured outputs serve as the foundation for diverse decision-making tasks, linking drawing-derived information to process planning, tool selection, inspection setup, and cost estimation. Effective integration also depends on bridging the perspectives of designers and manufacturers. Although this interpretive gap is beyond the scope of the present work, an essential prerequisite is addressed by converting information from 2D drawings into structured, machine-readable form. In this way, previously unstructured annotations are transformed into actionable inputs that enable rule-based systems to support end-to-end manufacturing operations.

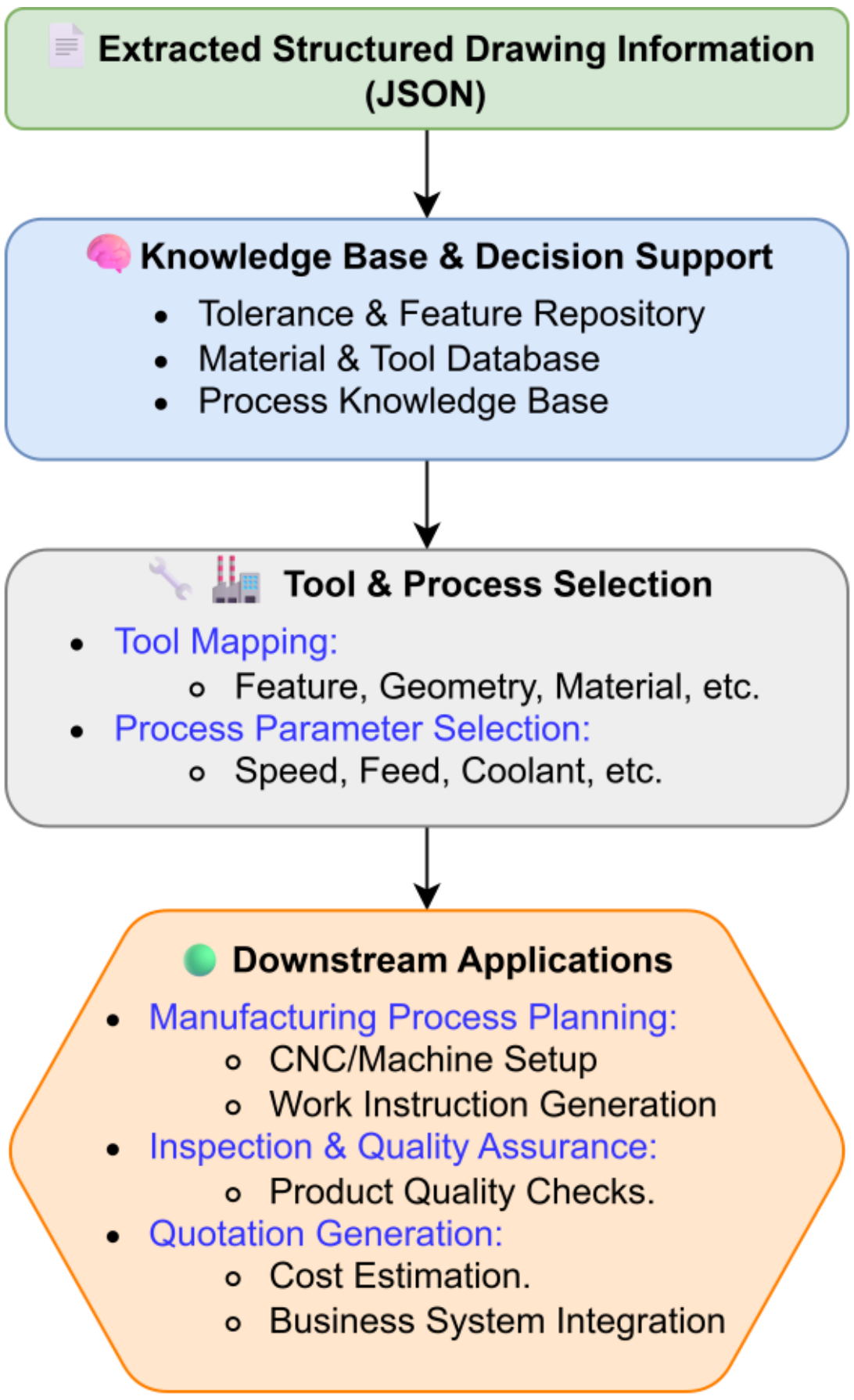

**Fig. 16**. Integration of extracted engineering drawing information into downstream manufacturing workflows.

To illustrate this capability, the following case study presents a real-world application using a shaft drawing. It demonstrates how the structured outputs extracted by the proposed method can be interpreted through a rule-based reasoning engine to inform machining decisions, select appropriate tools, and support planning activities consistent with model-based manufacturing principles.

## 5.1 Interpretation Preconditions

To ensure reliable interpretation, the framework operates under a set of key assumptions:

1. The part material (with properties) is either extracted from the drawing or supplied externally.
2. The drawing adheres to recognized standards (e.g., ASME Y14.5 [2] for GD&T and ISO 21920 [50] for surface finish).
3. All relevant annotations are accurately parsed into structured JSON, as shown in Fig. 16. Any misclassified or omitted annotation may lead to downstream errors (e.g. a missing tolerance might cause an improper process choice [29]).
4. The manufacturing environment has a defined set of allowable tools and operations, maintained in a local tooling database.

These preconditions ensure semantic completeness, prevent misclassification-driven errors, and align the interpretation process with real-world constraints such as tooling availability and compliance requirements.

### 5.2 Rule-Based Reasoning Engine

At the core of the reasoning framework lies a deterministic rule engine grounded in KBE principles. It evaluates annotation properties such as size, tolerance class, surface finish, and material to infer a corresponding set of machining operations, tools, and parameters. Each rule is crafted based on domain standards including, ASME Y14.5, ISO 21920, ISO IT grade tables [51], and supplemented by industrial best practices drawn from references such as the Machinery's Handbook [52] and vendor-specific guidelines (e.g., Sandvik Coromant). These rules encompass both formal knowledge and expert heuristics into traceable, machine-interpretable logic.

The parsed output from the drawing under study as shown in Fig. 16 is structured into the standardized schema introduced earlier. This structured representation serves as direct input to the rule-based reasoning system, which interprets the structured data to recommend suitable manufacturing actions.

To illustrate the core inference process, a set of representative rules is provided in Table 6, capturing deterministic mappings between feature-level inputs and machining recommendations. Each rule defines a set of input conditions, such as feature type, dimensions, tolerances, and material properties, and maps them to a recommended set of operations and compatible tools. These mappings are aligned with international standards and validated shop-floor heuristics to ensure industrial relevance and traceability.

**Table 6.** Representative rule table capturing feature-based interpretation logic.

| Feature Type | Conditions | Recommended Operations | Tool Selection |
|---|---|---|---|
| Threaded Hole | 6×M5 TAP THROUGH; Material: C-45 Steel | Drill Ø4.2 mm, Tap M5×0.8 | Twist Drill (Ø4.2 mm, HSS); Spiral Flute Tap (M5×0.8 mm, HSS-E, TiN-coated) |
| Cylindrical Shaft | Ø28 mm; Tolerance ±0.05 mm; Surface Roughness Ra 0.8 μm | Rough Turning, Finish Turning | Carbide Insert (Roughing); Wiper Insert (Finishing) |
| Hole (Tight Tolerance) | Diameter Tolerance within ISO IT6–IT8 | Drill + Finish Operation (Reaming/Boring) | Reamer (H7 grade); Precision Drill |
| Shaft (Ultra-Fine Finish) | Surface Roughness Ra < 0.4 μm or tight diameter tolerance | Fine Turning or Grinding | Super-finishing Insert: Wiper Insert; Grinding Wheel |
| Positional Feature | Ø0.020 mm Positional Tolerance applied to Datums A, B, and C | CNC Finishing with In-Process Probing | Custom Fixturing Setup; CNC Touch Probe; Adaptive Pathing |
| Hole or Shaft | Loose Tolerance (e.g., ISO IT12) | Single Roughing Pass | Standard Carbide Insert |

For instance, when the system detects a threaded hole annotated as "6×M5 TAP THROUGH" in a C-45 steel component, it automatically infers:

- **Operations:** Drill Ø4.2 mm, Tap M5×0.8 mm

- **Tool Selections:** Twist Drill (Ø4.2 mm, HSS), Spiral Flute Tap (M5×0.8 mm, HSS-E, TiN-coated)

This rule reflects ISO thread standards [53] for M5×0.8 mm pitch in medium-carbon steel, ensuring optimal engagement (~75%) and tool-material compatibility. In another case, the detection of a 28 mm diameter shaft with a ±0.05 mm tolerance and surface roughness Ra 0.8 μm leads to:

- **Operations:** Rough Turning, Finish Turning
- **Tool Selections:** Carbide Insert (Roughing), Wiper Insert (Finishing)

The use of wiper inserts enables achieving Ra ≤ 0.8 μm at higher feed rates, aligning with modern finishing strategies in turning operations [54]. Other rule examples address feature-specific cases such as ultra-fine surface finishes, tight tolerance ranges (IT6–IT8), loose tolerances (e.g., IT12), and high-precision positional tolerancing relative to datum references. These rules are invoked dynamically based on the extracted annotation values and drive the downstream manufacturing process and tool selection. By encoding expert decision logic into explicit rule sets, the system produces traceable, explainable outcomes that are essential for bridging CAD/CAM automation with human oversight in industrial contexts.

### 5.3 Tool and Parameter Database

To operationalize rule outputs, a structured tooling and process parameter database is used to retrieve compatible tools and machining parameters. This database is organized according to ISO 13399 [55] to ensure interoperability with CAM systems and vendor catalogs. It stores geometric and material attributes for each tool, linked with cutting speeds and feed values for specific material-tool combinations. The system queries this database by filtering for compatibility criteria and retrieving optimal parameters for process planning. Sample tool entries, schema structure, and query examples are provided in **Appendix B**, demonstrating how the database acts as a digital library for real-time lookup and integration.

### 5.4 Downstream Process Integration

Finally, the output of the rule engine is integrated into a downstream manufacturing planning framework to support comprehensive decision-making. As illustrated in Fig. 17, the structured annotations enable feature-based reasoning, guiding operation sequencing, machining parameter selection, and inspection planning. These outputs enable alignment with manufacturing execution systems (MES), quotation workflows, and quality assurance procedures, bridging the gap between engineering drawings and executable production strategies.

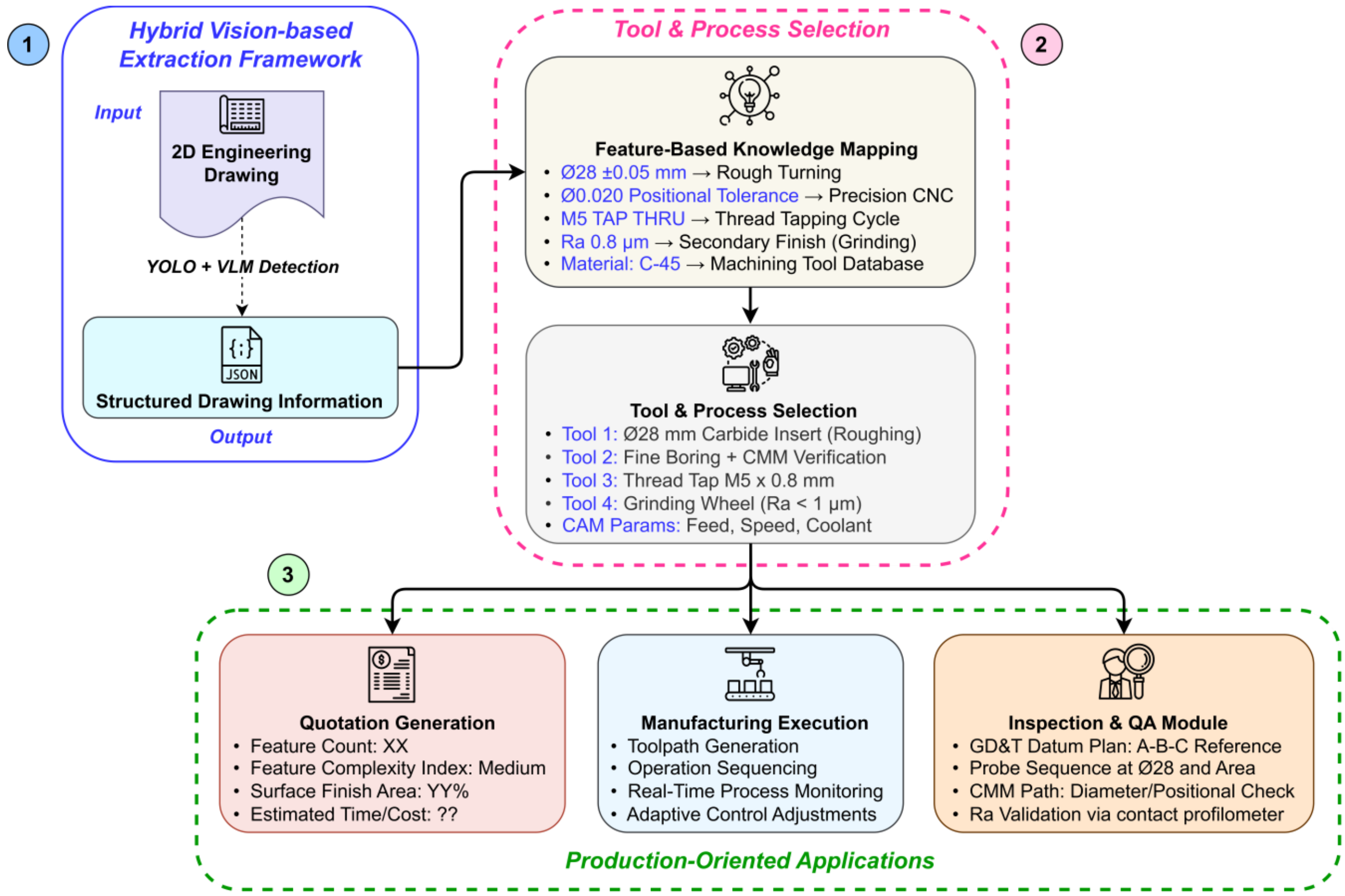

**Fig. 17.** Schematic overview of the rule-driven interpretation pipeline, linking structured drawing information to downstream planning and manufacturing decision support.

This case study illustrates the broader value of the proposed system. By extracting semantically structured data from traditional drawings and embedding it into a formal decision-making framework, the pipeline delivers automation and consistency, which are key enablers for intelligent and scalable manufacturing workflows.

## 6. Conclusions

This work presented a hybrid vision-based framework for the automated extraction of structured information from 2D engineering drawings, a longstanding bottleneck in digital manufacturing workflows. By combining a rotation-aware object detector with a lightweight VLM, the proposed pipeline effectively localized and parsed diverse annotation types into structured data formats. This effort was supported by a manually curated dataset of 1,367 engineering drawings spanning nine industrially relevant categories, enabling both model training and benchmarking.

The main contribution is a two-stage hybrid framework that combines YOLOv11-obb with fine-tuned vision-language parsing to extract structured information from engineering drawings. This was supported by an annotated drawing dataset and a comparative evaluation of two fine-tuned VLMs (Donut and Florence-2) against manually verified ground truth across four key evaluation metrics. The proposed pipeline demonstrated reliable localization and parsing of various annotation categories into structured outputs. Empirical results showed strong performance, with Donut achieving high recall (99.2%) and balanced precision and F1-score, while maintaining a low hallucination rate. These results underscored the framework's effectiveness in modernizing drawing interpretation and enhancing automation and data interoperability in digital manufacturing workflows. Although cascading errors could be a potential concern, the two-stage design that separates spatial detection from semantic parsing helped to mitigate their impact. The high accuracy of detection provided a strong foundation for reliable downstream parsing. A case study further demonstrated the practical utility of the extracted outputs in downstream tasks such as tool and process selection.

However, the study has several limitations. Despite careful curation, the dataset remains limited in scope and may not fully capture the diversity and complexity of real-world industrial drawings, particularly those involving large assemblies or non-standard annotation formats. Although the dataset spans multiple domains (aerospace, automotive, and general mechanical engineering) and includes both part-level and assembly-level drawings across CAD-derived and scanned formats, further expansion remains necessary to capture broader industrial diversity. Moreover, the downstream reasoning in the case study assumes ideal conditions and standardized tooling, which may restrict generalizability to broader industrial contexts. Nevertheless, the case study demonstrates that the extracted structured information is both accurate and actionable for downstream tasks such as tool and process selection. Broader validation across multiple manufacturing domains (e.g., machining, additive manufacturing, sheet metal, inspection, etc.), along with integration into CAD/CAM environments where structured outputs can drive real digital workflows will be pursued in future work to demonstrate wider industrial relevance.

Future work will aim to address the current dataset's limited coverage by expanding it to include more varied and complex engineering drawings, with improved representation of underrepresented annotation types. In addition, imbalance-aware strategies such as focal loss, class-weighted training, and sophisticated resampling techniques can be explored to further strengthen robustness for rare annotation categories. Industrial-scale validation will also be pursued by deploying the structured outputs in CAD/CAM workflows and inspection pipelines to evaluate applicability in real manufacturing settings. Future work will also include a systematic study of potential cascading error effects, by benchmarking parsing performance on ground truth and detected patches, to provide a more detailed understanding of how detection quality influences downstream semantic extraction. Efforts will also focus on incorporating mechanisms for error detection and correction to minimize manual intervention. Furthermore, integrating the structured outputs with real-time manufacturing feedback and developing adaptive, context-aware decision-making modules will enhance the framework's robustness and applicability in diverse production environments.

## Declaration of Competing Interest

The authors declare that they have no known competing financial interests or personal relationships that could have appeared to influence the work reported in this paper.

## Acknowledgement

This work is supported by the Agency for Science, Technology and Research (A*STAR), Singapore, through the RIE2025 MTC IAF-PP grant (Grant No. M22K5a0045). It is also supported by Singapore International Graduate Award (SINGA) (Awardee: Muhammad Tayyab Khan) funded by A*STAR and Nanyang Technological University, Singapore.

## References

[1] Y.-H. Lin, Y.-H. Ting, Y.-C. Huang, K.-L. Cheng, and W.-R. Jong, "Integration of Deep Learning for Automatic Recognition of 2D Engineering Drawings," *Machines*, vol. 11, no. 8, Art. no. 8, Aug. 2023, doi: 10.3390/machines11080802.

[2] "Y14.5 Dimensioning and Tolerancing - ASME." Accessed: Sep. 27, 2024. [Online]. Available: https://www.asme.org/codes-standards/find-codes-standards/y14-5-dimensioning-tolerancing

[3] W. Sun and Y. Gao, "A datum-based model for practicing geometric dimensioning and tolerancing," *J. Eng. Technol.*, vol. 35, pp. 38–47, Sep. 2018.

[4] "AutoCAD Mechanical 2022 Help | About Balloons (AutoCAD Mechanical Toolset) | Autodesk." Accessed: Sep. 27, 2024. [Online]. Available: https://help.autodesk.com/view/AMECH_PP/2022/ENU/?guid=GUID-F12F0EA0-0810-42EE-A3FE-327041AFAEEE

[5] "Data Management and SPC Software," MeasurLink. Accessed: Sep. 27, 2024. [Online]. Available: https://measurlink.com/

[6] J. Redmon, S. Divvala, R. Girshick, and A. Farhadi, "You Only Look Once: Unified, Real-Time Object Detection," in *2016 IEEE Conference on Computer Vision and Pattern Recognition (CVPR)*, Las Vegas, NV, USA: IEEE, Jun. 2016, pp. 779–788. doi: 10.1109/CVPR.2016.91.

[7] M. T. Khan, L. Chen, Y. H. Ng, W. Feng, N. Y. J. Tan, and S. K. Moon, "Fine-Tuning Vision-Language Model for Automated Engineering Drawing Information Extraction," Nov. 06, 2024, *arXiv*: arXiv:2411.03707. doi: 10.48550/arXiv.2411.03707.

[8] J. V. Toro and M. Tarkian, "Optimizing Text Recognition in Mechanical Drawings: A Comprehensive Approach," *Machines*, vol. 13, no. 3, p. 254, Mar. 2025, doi: 10.3390/machines13030254.

[9] "Oriented Bounding Boxes Object Detection - Ultralytics YOLO Docs." Accessed: Apr. 18, 2025. [Online]. Available: https://docs.ultralytics.com/tasks/obb/

[10] G. Kim *et al.*, "OCR-Free Document Understanding Transformer," in *Computer Vision – ECCV 2022*, vol. 13688, S. Avidan, G. Brostow, M. Cissé, G. M. Farinella, and T. Hassner, Eds., in Lecture Notes in Computer Science, vol. 13688. , Cham: Springer Nature Switzerland, 2022, pp. 498–517. doi: 10.1007/978-3-031-19815-1_29.

[11] B. Xiao *et al.*, "Florence-2: Advancing a Unified Representation for a Variety of Vision Tasks," in *2024 IEEE/CVF Conference on Computer Vision and Pattern Recognition (CVPR)*, Seattle, WA, USA: IEEE, Jun. 2024, pp. 4818–4829. doi: 10.1109/CVPR52733.2024.00461.

[12] A. Corallo, V. Del Vecchio, M. Lezzi, and A. Luperto, "Model-Based Enterprise Approach in the Product Lifecycle Management: State-of-the-Art and Future Research Directions," *Sustainability*, vol. 14, no. 3, Art. no. 3, Jan. 2022, doi: 10.3390/su14031370.

[13] M. Dalibor, N. Jansen, B. Rumpe, L. Wachtmeister, and A. Wortmann, "Model-Driven Systems Engineering for Virtual Product Design," in *2019 ACM/IEEE 22nd International Conference on Model Driven Engineering Languages and Systems Companion (MODELS-C)*, Munich, Germany: IEEE, Sep. 2019, pp. 431–436. doi: 10.1109/MODELS-C.2019.00069.

[14] W. Khallouli, M. S. Uddin, A. Sousa-Poza, J. Li, and S. Kovacic, “Leveraging Transformer-Based OCR Model with Generative Data Augmentation for Engineering Document Recognition,” *Electronics*, vol. 14, no. 1, Art. no. 1, Jan. 2025, doi: 10.3390/electronics14010005.

[15] J. Gao, D. T. Zheng, N. Gindy, and D. Clark, “Extraction/conversion of geometric dimensions and tolerances for machining features,” *Int. J. Adv. Manuf. Technol.*, vol. 26, no. 4, pp. 405–414, Aug. 2005, doi: 10.1007/s00170-004-2195-3.

[16] D. B. Lysak, P. M. Devaux, and R. Kasturi, “View labeling for automated interpretation of engineering drawings,” *Pattern Recognit.*, vol. 28, no. 3, pp. 393–407, Mar. 1995, doi: 10.1016/0031-3203(94)00103-S.

[17] L. Jamieson, C. Francisco Moreno-García, and E. Elyan, “A review of deep learning methods for digitisation of complex documents and engineering diagrams,” *Artif. Intell. Rev.*, vol. 57, no. 6, p. 136, May 2024, doi: 10.1007/s10462-024-10779-2.

[18] Y. Xu *et al.*, “Tolerance Information Extraction for Mechanical Engineering Drawings - A Digital Image Processing and Deep Learning-based Model,” *CIRP J. Manuf. Sci. Technol.*, vol. 50, pp. 55–64, Mar. 2024, doi: 10.1016/j.cirpj.2024.01.013.

[19] M. Francois, V. Eglin, and M. Biou, “Text Detection and Post-OCR Correction in Engineering Documents,” 2022, pp. 726–740. doi: 10.1007/978-3-031-06555-2_49.

[20] S. Mani, M. A. Haddad, D. Constantini, W. Douhard, Q. Li, and L. Poirier, “Automatic Digitization of Engineering Diagrams using Deep Learning and Graph Search,” in *2020 IEEE/CVF Conference on Computer Vision and Pattern Recognition Workshops (CVPRW)*, Seattle, WA, USA: IEEE, Jun. 2020, pp. 673–679. doi: 10.1109/CVPRW50498.2020.00096.

[21] E.-S. Yu, J.-M. Cha, T. Lee, J. Kim, and D. Mun, “Features Recognition from Piping and Instrumentation Diagrams in Image Format Using a Deep Learning Network,” *Energies*, vol. 12, no. 23, Art. no. 23, Jan. 2019, doi: 10.3390/en12234425.

[22] M. S. M. Yazed, E. F. A. Shaubari, and M. H. Yap, “A Review of Neural Network Approach on Engineering Drawing Recognition and Future Directions,” *JOIV Int. J. Inform. Vis.*, vol. 7, no. 4, pp. 2513–2522, Dec. 2023, doi: 10.62527/joiv.7.4.1716.

[23] “High QA - Manufacturing Quality Experts.” Accessed: Apr. 13, 2025. [Online]. Available: https://www.highqa.com/

[24] B. Faltin, P. Schönfelder, and M. König, “Improving Symbol Detection on Engineering Drawings Using a Keypoint-Based Deep Learning Approach,” *EG-ICE 20223 Conf. Pap.*, pp. 1–10, Jul. 2023.

[25] M. T. Khan, L. Chen, Y. H. Ng, W. Feng, N. Y. J. Tan, and S. K. Moon, “Leveraging Vision-Language Models for Manufacturing Feature Recognition in Computer-Aided Designs,” *J. Comput. Inf. Sci. Eng.*, vol. 25, no. 104501, Aug. 2025, doi: 10.1115/1.4069266.

[26] Y. Xu, M. Li, L. Cui, S. Huang, F. Wei, and M. Zhou, “LayoutLM: Pre-training of Text and Layout for Document Image Understanding,” in *Proceedings of the 26th ACM SIGKDD International Conference on Knowledge Discovery & Data Mining*, Aug. 2020, pp. 1192–1200. doi: 10.1145/3394486.3403172.

[27] S. Appalaraju, B. Jasani, B. U. Kota, Y. Xie, and R. Manmatha, “DocFormer: End-to-End Transformer for Document Understanding,” in *2021 IEEE/CVF International Conference on Computer Vision (ICCV)*, Montreal, QC, Canada: IEEE, Oct. 2021, pp. 973–983. doi: 10.1109/ICCV48922.2021.00103.

[28] C. Gu, K. Lin, Y. Luo, J. Hou, and X.-Y. Li, “ViRED: Prediction of Visual Relations in Engineering Drawings,” presented at the 2024 20th International Conference on Mobility, Sensing and Networking (MSN), IEEE Computer Society, Dec. 2024, pp. 681–688. doi: 10.1109/MSN63567.2024.00097.

[29] L. Xie *et al.*, “Graph neural network-enabled manufacturing method classification from engineering drawings,” *Comput. Ind.*, vol. 142, p. 103697, Nov. 2022, doi: 10.1016/j.compind.2022.103697.

[30] J. Gao, D. T. Zheng, N. Gindy, and D. Clark, “Extraction/conversion of geometric dimensions and

tolerances for machining features," *Int. J. Adv. Manuf. Technol.*, vol. 26, no. 4, pp. 405–414, Aug. 2005, doi: 10.1007/s00170-004-2195-3.

[31] R. Dzhusupova, R. Banotra, J. Bosch, and H. H. Olsson, "Pattern Recognition Method for Detecting Engineering Errors on Technical Drawings," in *2022 IEEE World AI IoT Congress (AIIoT)*, Jun. 2022, pp. 642–648. doi: 10.1109/AIIoT54504.2022.9817294.

[32] J. Esanakula, ·Naga venkata Sridhar, and V. Rangadu, "Knowledge Based Engineering: Notion, Approaches and Future Trends," *Am. J. Intell. Syst.*, vol. 2015, pp. 1–17, Jan. 2015, doi: 10.5923/j.ajis.20150501.01.

[33] G. Stavropoulou *et al.*, "Digital Twin Meets Knowledge Graph for Intelligent Manufacturing Processes," *Sensors*, vol. 24, no. 8, p. 2618, Jan. 2024, doi: 10.3390/s24082618.

[34] "Leading Image & Video Data Annotation Platform | CVAT." Accessed: Mar. 22, 2025. [Online]. Available: https://www.cvat.ai

[35] P. Company, R. Plumed, C. González-Lluch, M. Contero, and J. D. Camba, "A strategy to classify and resolve ambiguities when reverse engineering drawings to CAD models," *Cogent Eng.*, vol. 12, no. 1, p. 2438810, Dec. 2025, doi: 10.1080/23311916.2024.2438810.

[36] X. Xie, G. Cheng, J. Wang, X. Yao, and J. Han, "Oriented R-CNN for Object Detection," in *2021 IEEE/CVF International Conference on Computer Vision (ICCV)*, Montreal, QC, Canada: IEEE, Oct. 2021, pp. 3500–3509. doi: 10.1109/ICCV48922.2021.00350.

[37] J. Han, J. Ding, N. Xue, and G.-S. Xia, "ReDet: A Rotation-equivariant Detector for Aerial Object Detection," in *2021 IEEE/CVF Conference on Computer Vision and Pattern Recognition (CVPR)*, Nashville, TN, USA: IEEE, Jun. 2021, pp. 2785–2794. doi: 10.1109/CVPR46437.2021.00281.

[38] Ultralytics, "OBB." Accessed: Sep. 02, 2025. [Online]. Available: https://docs.ultralytics.com/tasks/obb

[39] "COCO Dataset - Ultralytics YOLO Docs." Accessed: Apr. 17, 2025. [Online]. Available: https://docs.ultralytics.com/datasets/detect/coco/#usage

[40] "The ASME Y14.5 GD&T Standard | GD&T Basics." Accessed: Apr. 19, 2025. [Online]. Available: https://www.gdandtbasics.com/asme-y14-5-gdt-standard/

[41] "torchvision.transforms — Torchvision master documentation." Accessed: Mar. 22, 2025. [Online]. Available: https://pytorch.org/vision/0.9/transforms.html

[42] "naver-clova-ix/donut-base · Hugging Face." Accessed: Jun. 09, 2025. [Online]. Available: https://huggingface.co/naver-clova-ix/donut-base

[43] Z. Liu *et al.*, "Swin Transformer: Hierarchical Vision Transformer using Shifted Windows," in *2021 IEEE/CVF International Conference on Computer Vision (ICCV)*, Montreal, QC, Canada: IEEE, Oct. 2021, pp. 9992–10002. doi: 10.1109/ICCV48922.2021.00986.

[44] Y. Liu *et al.*, "Multilingual Denoising Pre-training for Neural Machine Translation," *Trans. Assoc. Comput. Linguist.*, vol. 8, pp. 726–742, 2020, doi: 10.1162/tacl_a_00343.

[45] M. T. Khan, Z. Yong, L. Chen, J. M. Tan, W. Feng, and S. K. Moon, "Automated Parsing of Engineering Drawings for Structured Information Extraction Using a Fine-tuned Document Understanding Transformer," May 02, 2025, *arXiv*: arXiv:2505.01530. doi: 10.48550/arXiv.2505.01530.

[46] "microsoft/Florence-2-base · Hugging Face." Accessed: Jun. 09, 2025. [Online]. Available: https://huggingface.co/microsoft/Florence-2-base

[47] M. Ding, B. Xiao, N. Codella, P. Luo, J. Wang, and L. Yuan, "DaViT: Dual Attention Vision Transformers," in *Computer Vision – ECCV 2022*, vol. 13684, S. Avidan, G. Brostow, M. Cissé, G. M. Farinella, and T. Hassner, Eds., in Lecture Notes in Computer Science, vol. 13684. , Cham: Springer Nature Switzerland, 2022, pp. 74–92. doi: 10.1007/978-3-031-20053-3_5.

[48] T. Chen, S. Saxena, L. Li, D. J. Fleet, and G. Hinton, "Pix2seq: A Language Modeling Framework for

Object Detection," Mar. 27, 2022, *arXiv*: arXiv:2109.10852. doi: 10.48550/arXiv.2109.10852.

[49] K. Lv, Y. Yang, T. Liu, Q. Guo, and X. Qiu, "Full Parameter Fine-tuning for Large Language Models with Limited Resources," in *Proceedings of the 62nd Annual Meeting of the Association for Computational Linguistics (Volume 1: Long Papers)*, L.-W. Ku, A. Martins, and V. Srikumar, Eds., Bangkok, Thailand: Association for Computational Linguistics, Aug. 2024, pp. 8187–8198. doi: 10.18653/v1/2024.acl-long.445.

[50] "ISO 21920-1:2021," ISO. Accessed: May 14, 2025. [Online]. Available: https://www.iso.org/standard/72196.html

[51] "International Tolerance (IT) Grades ISO 286-1 - 2010(E) Table Chart." Accessed: May 14, 2025. [Online]. Available: https://www.engineersedge.com/international_tol.htm#google_vignette

[52] "Machinery's Handbook 29th Edition." Accessed: May 14, 2025. [Online]. Available: https://dl.icdst.org/pdfs/files4/80364b03673ba30eb5ccf1e27e119ffc.pdf

[53] "ISO 529:2017 - Short machine taps and hand taps," iTeh Standards. Accessed: May 15, 2025. [Online]. Available: https://standards.iteh.ai/catalog/standards/iso/2a88e2d4-ca61-475e-a535-80496ea77a81/iso-529-2017

[54] Caro, "Surface Roughness Explained: Ra, Rq, Rz, and More," Richconn | Precision CNC Parts Manufacturing | China CNC Machining Manufacturer. Accessed: May 15, 2025. [Online]. Available: https://richconn.com/surface-roughness/

[55] "ISO 13399-1:2006," ISO. Accessed: Aug. 28, 2025. [Online]. Available: https://www.iso.org/standard/36757.html

# Appendices

## Appendix A: Additional Annotation Parsing Results Demonstrated via In-House GUI

This appendix presents additional parsing results generated by the proposed framework, visualized through the in-house developed GUI. For each example, the original input drawing, overlaid annotation regions with confidence scores, and the corresponding structured JSON outputs are included. These examples illustrate the model's ability to generalize across diverse annotation styles, drawing formats, and complexity levels. Collectively, they reinforce the robustness of the proposed pipeline and demonstrate its practical applicability to manufacturing documentation and CAD/CAM integration workflows.

### Example 1

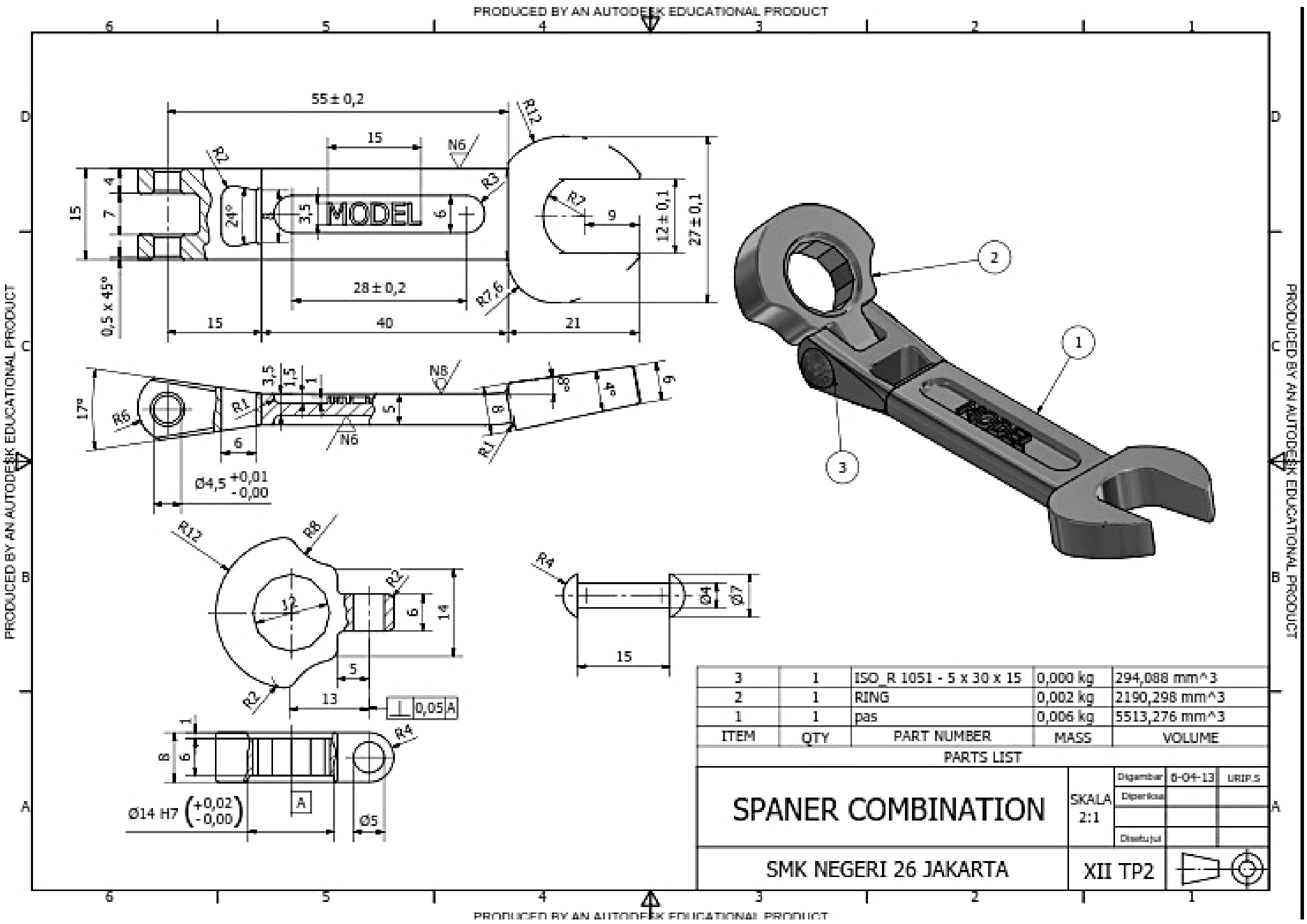

**Fig. A1.** Original 2D engineering drawing: *Spaner Combination.*

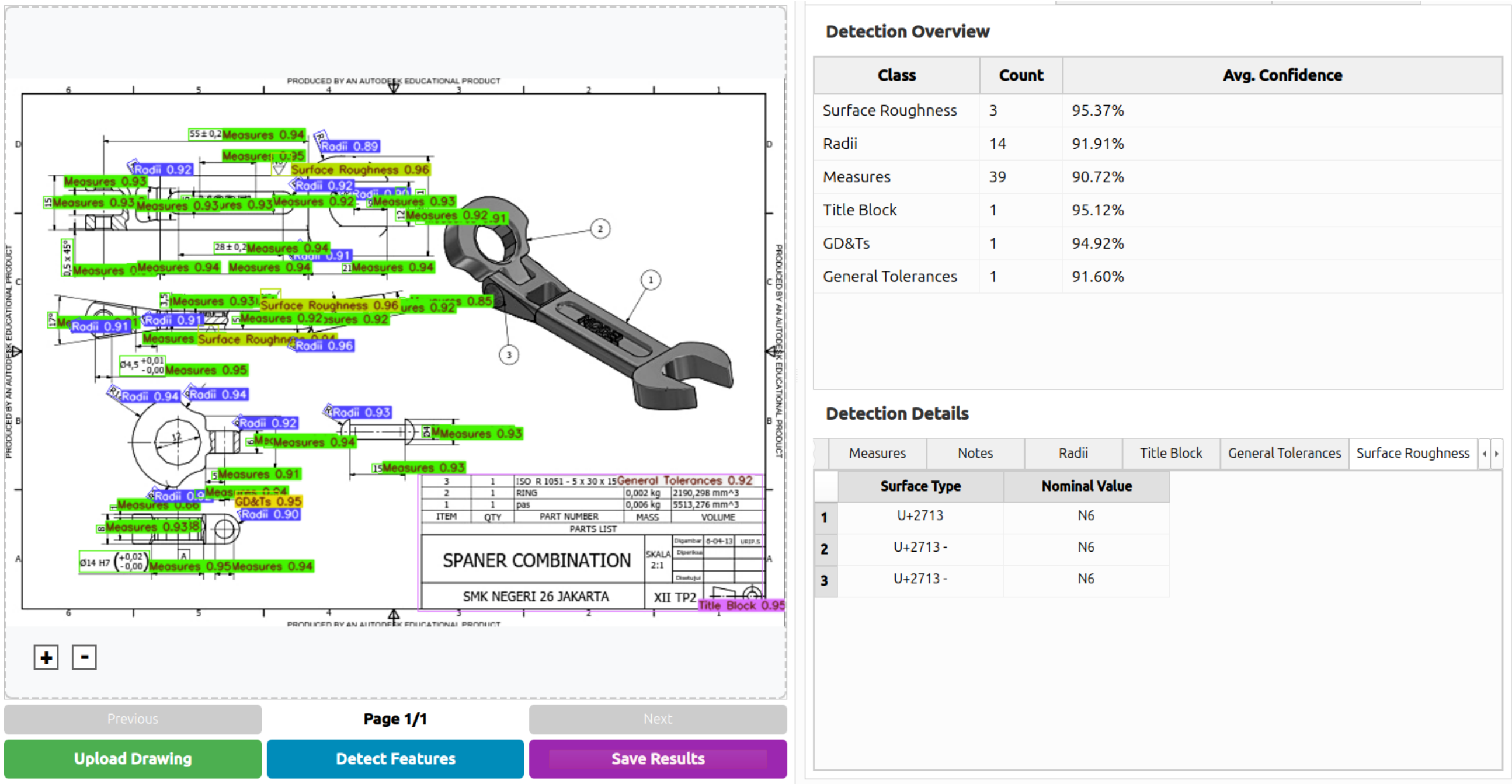

**Fig. A2.** Detected annotation regions overlaid on the drawing using YOLOv11-obb.

```
{
"Material": "",
"Threads": “",
"GD&T": [
   {"Type": "Perpendicularity", "Tolerance": "0.05", "Datums": ["A"]}],
"General Tolerance": "ISO_R 1051 – 5 x 30 x 15",
“Radii”: {“R1”, “R2”, “R3”, “R4”, “R6”, “R7”, “R8”, “R12”},
"Surface Roughness": [
   {"Ra": "N6"},
   {"Ra": "N8"}],
"Measures": [
   {"Feature": "Overall Length", "Value": "55 ±0.2 mm"},
   {"Feature": "Jaw Width", "Value": "7.2 ±0.1 mm"},
   {"Feature": "Handle Thickness", "Value": "9.5 ±0.4 mm"},
   {"Feature": "Hole Diameter", "Value": "Ø5 +0.01/-0.00 mm"},
   {"Feature": "Ring Width", "Value": "Ø14 H7 (+0.02/-0.00 mm)"},
   {"Feature": "Slot Length", "Value": "28 ±0.2 mm"},
   {"Feature": "Opening Width", "Value": "13 mm"}],
"Title Block": {
   "Designer": "SMK NEGERI 26 JAKARTA",
   "Date": “6-04-13",
   “Scale”: “2:1”
   "Drawing Name": "Spaner Combination"},
 "Notes": “”
}
```

**Fig. A3.** Structured JSON output generated via vision-language parsing of the localized regions.

## Example 2

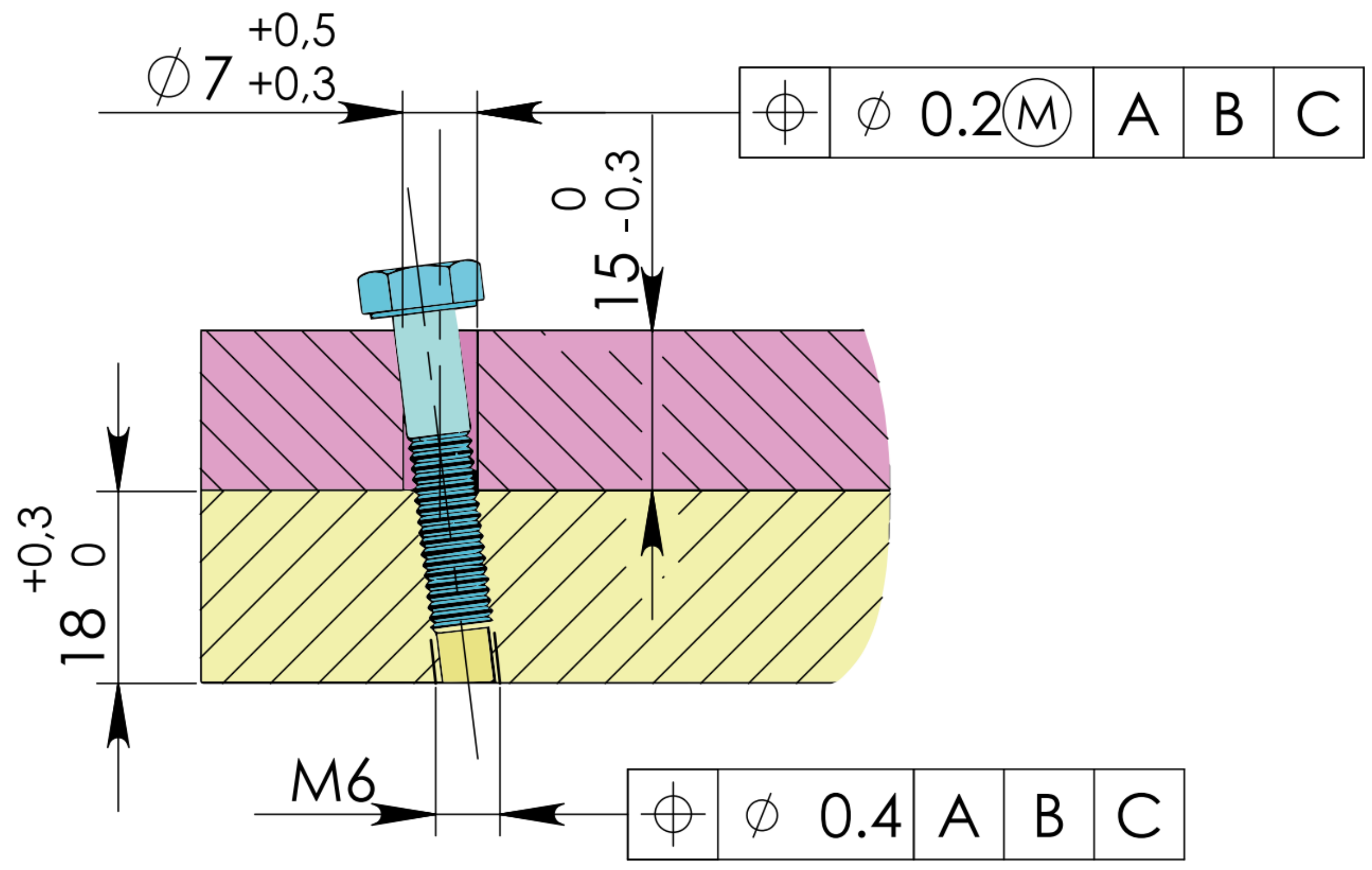

**Fig. A4.** Original 2D engineering drawing.

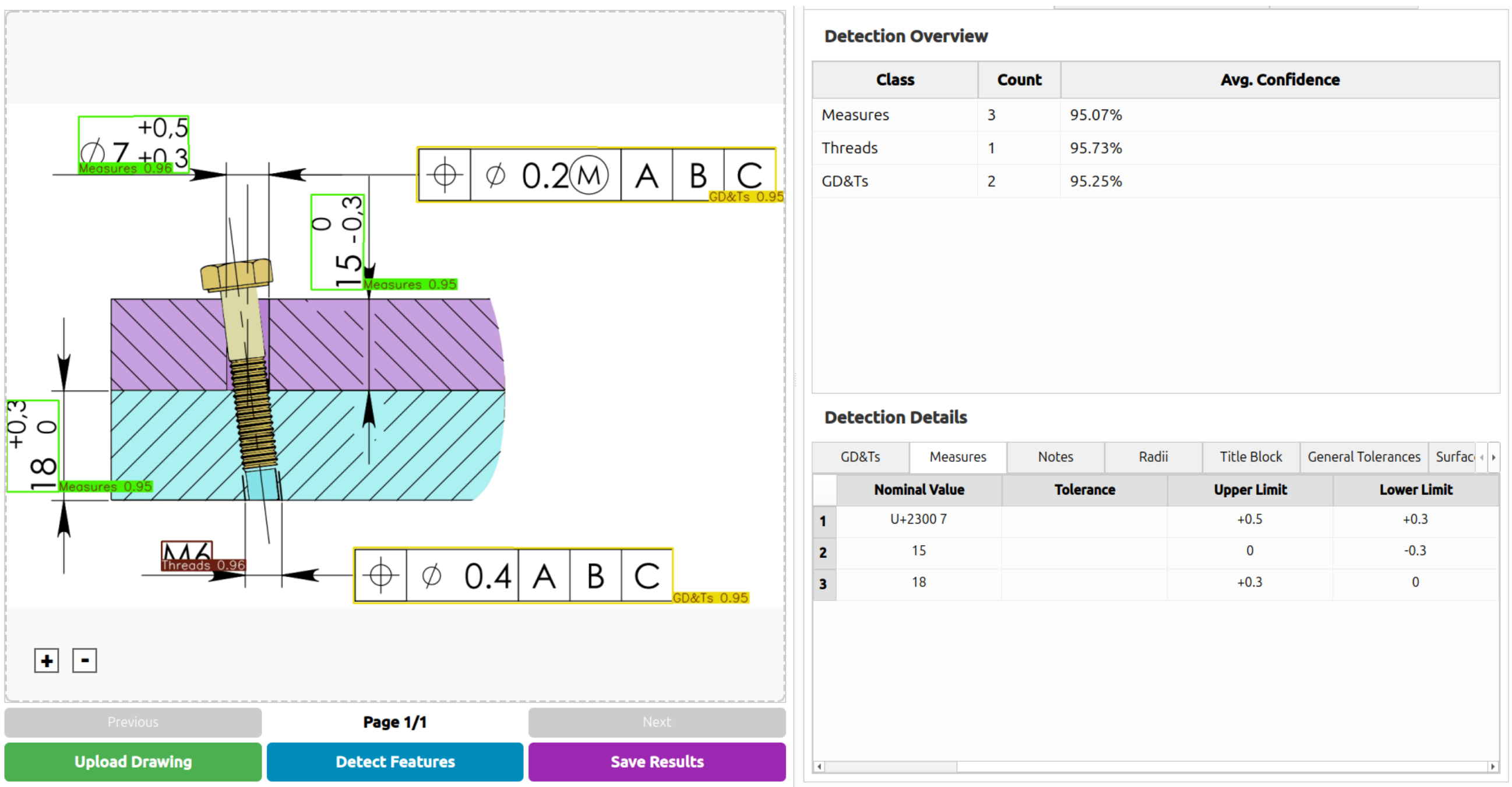

**Fig. A5.** YOLOv11-obb-based detection overlay showing oriented annotation regions.

```
{
"Material": "",
"Threads": [{"Type": "M6"}],
"GD&T": [
  {"Type": "Position", "Tolerance": "Ø0.2 (M)", "Datums": ["A", "B", "C"]},
  {"Type": "Position", "Tolerance": "Ø0.4", "Datums": ["A", "B", "C"]}],
"General Tolerance": "",
"Radii": "",
"Surface Roughness": "",
"Measures": [
  {"Feature": "Hole Diameter", "Value": "Ø7 +0.5/+0.3"},
  {"Feature": "Height", "Value": "18 +0.3/0"},
  {"Feature": "Threaded Depth", "Value": "15 0/-0.3"}],
"Title Block": "",
"Notes": ""
}
```

**Fig. A6.** Corresponding structured JSON output generated by the VLM parser.

## Example 3

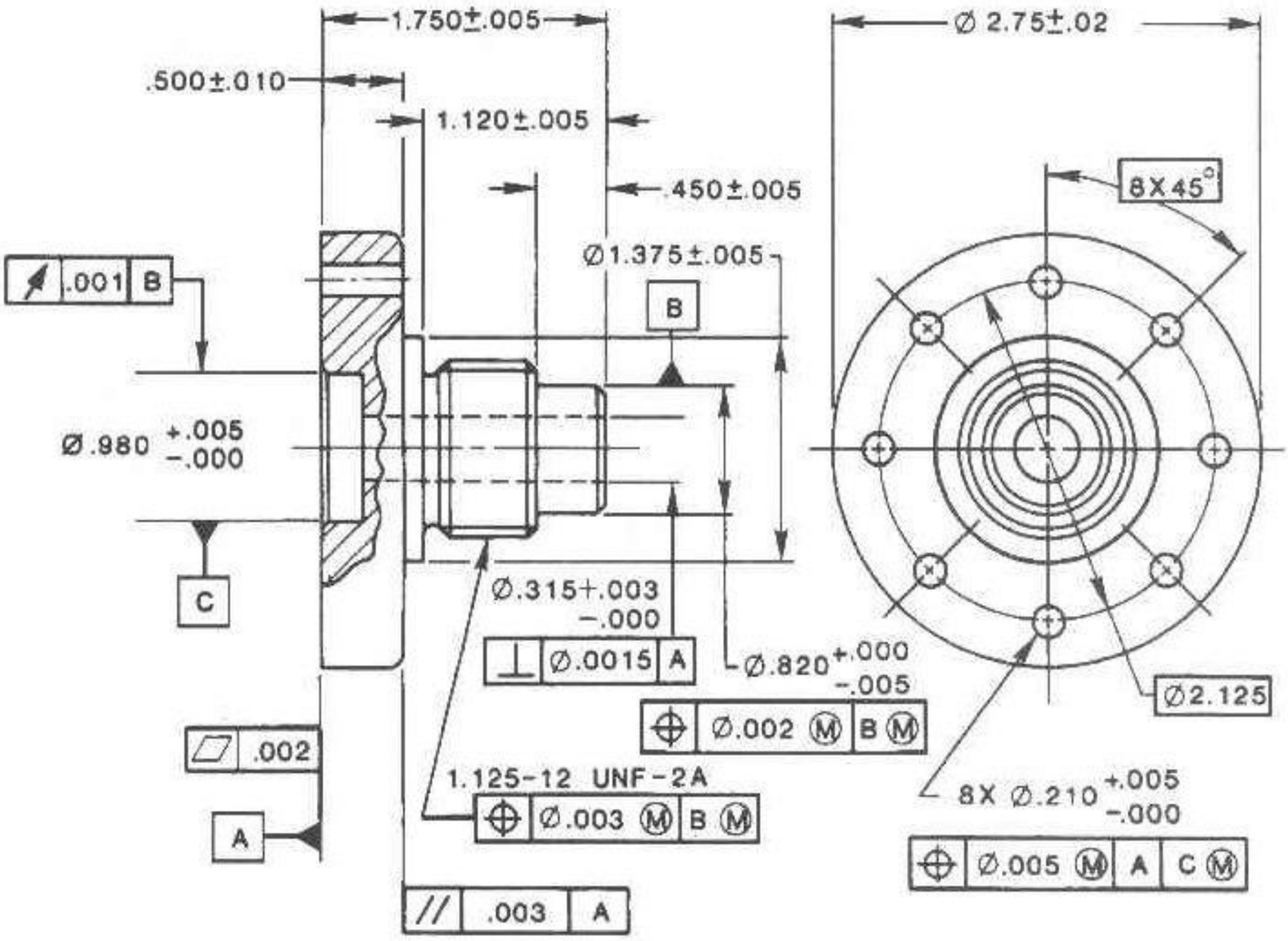

**Fig. A7.** Original 2D engineering drawing.

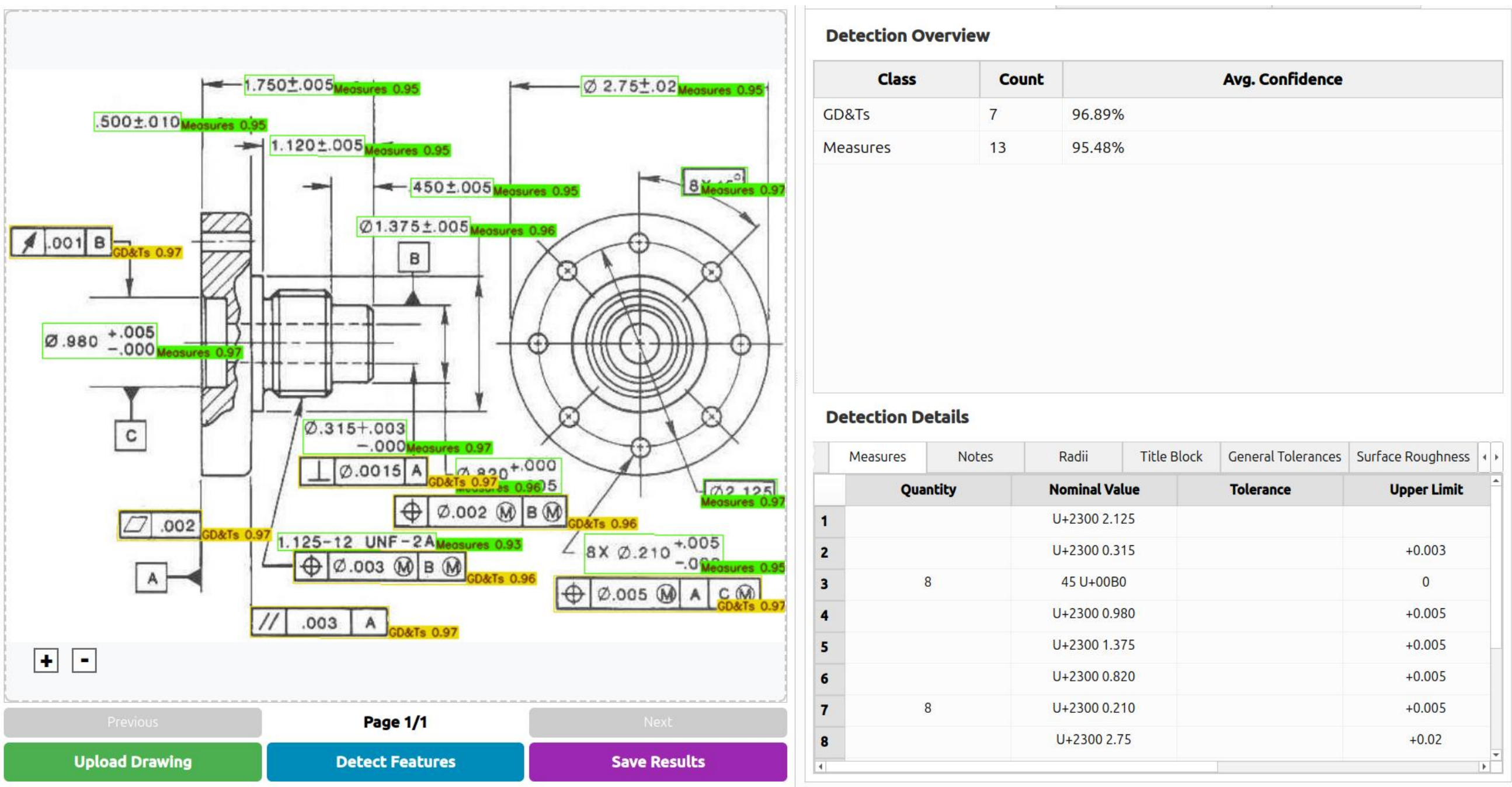

**Fig. A8.** Annotation localization using YOLOv11-obb with rotation-aware bounding boxes.

```
{
"Material": "",
"Threads": “”,
"GD&T": [
   {"Type": "Position", "Tolerance": "Ø0.002 (M)", "Datums": ["B (M)”]},
   {"Type": "Position", "Tolerance": "Ø0.003 (M)", "Datums": ["B (M)”]},
   {"Type": "Position", "Tolerance": "Ø0.005 (M)", "Datums": ["A", “C (M)”]},
   {"Type": "Flatness", "Tolerance": "0.002", "Datums": []},
   {"Type": "Perpendicularity", "Tolerance": "Ø0.0015", "Datums": ["A”]},
   {"Type": "Parallelism", "Tolerance": "0.003", "Datums": ["A”]},
   {"Type": "Circular Turnout", "Tolerance": "0.001", "Datums": ["B”]}],
"General Tolerance": "",
“Radii”: “”,
"Surface Roughness": “”,
"Measures": [
   {"Feature": "Face Thickness", "Value": "0.500 ±0.010"},
   {"Feature": "Total Length", "Value": "2.75 ±0.02"},
   {"Feature": "First Step Length", "Value": "1.750 ±0.005"},
   {"Feature": "Second Step Length", "Value": "1.120 ±0.005"},
   {"Feature": "Third Step Length", "Value": "0.450 ±0.005"},
   {"Feature": "Large Diameter", "Value": "Ø1.375 ±0.005"},
   {"Feature": "Shoulder Diameter", "Value": "Ø0.980 +0.005/-0.000"},
   {"Feature": "Internal Step Diameter", "Value": "Ø0.315 +0.003/-0.000"},
   {"Feature": "Outer Shaft Diameter", "Value": "Ø0.820 +0.005/-0.000"},
   {"Feature": "Bolt Hole Circle Diameter", "Value": "Ø2.125"},
   {"Feature": "Bolt Hole Diameter", "Value": "Ø0.210 +0.005/-0.000"},
   {"Feature": "Chamfer", "Value": "8X 45°"}],
"Title Block": “”,
"Notes": “”
}
```

**Fig. A9.** Parsed JSON output representing detected annotation content.

## Appendix B: Tool and Parameter Database Schema

This appendix presents the underlying data schema that supports rule-based tool selection and machining parameter retrieval in the digital twin environment. The schema follows ISO 13399 standards, enabling consistent and interoperable representation of tool attributes for both internal reasoning and external system integration.

```
{
Tool ID: Unique identifier (e.g. DRILL_4.2_HSS),
Tool Type: Category of tool (e.g. Twist Drill, Endmill, Reamer, Turning Insert),
Geometry: Key dimensions (for a drill: diameter and tip angle; for a turning insert: nose radius, inscribed circle, etc.),
Material: Tool material or grade (e.g. HSS, Carbide P20 grade, CBN),
Applicable Material: Workpiece materials the tool is suited for (e.g. steel up to 45 HRC, aluminum, etc.),
Cutting Parameters: Recommended cutting speed (m/min) and feed (mm/rev or mm/tooth) for each applicable material class.
Standard Codes: Optional field linking to ISO 13399 parameter codes or tool catalogue numbers.
}
```

**Fig. B1.** Tool schema and lookup logic for rule-based tool and parameter selection.

The tooling database is structured into two interrelated tables:

1. **Tool Inventory Table:** Stores geometric and classification metadata.
2. **Cutting Data Table:** Stores process parameters for specific tool-material combinations.

**Table B1.** Tool inventory with geometry, material, and classification fields (ISO 13399-compliant).

| Tool ID | Tool Type | Diameter | Material | Coating | ISO Code |
|---|---|---|---|---|---|
| T042 | Twist Drill | 4.2 mm | HSS | — | ISO13399_001 |
| T043 | Spiral Tap | M5×0.8 | HSS-E | TiN | ISO13399_002 |

**Table B2.** Recommended cutting parameters for tool-material combinations.

| Tool ID | Material | Speed (m/min) | Feed (mm/rev) |
|---|---|---|---|
| T042 | C-45 Steel | 20 | 0.1 |
| T043 | C-45 Steel | 15 | 0.05 |

**Lookup Example (Case Study Reference):**

When the rule-based system identifies a machining need, such as tapping an M5 thread in C-45 steel (as demonstrated in Section 5.2), it queries the tool database. From Table B1, it selects Tool T043, a spiral tap sized M5×0.8 suitable for the given material. Then, using Table B2, it retrieves the recommended cutting speed of 15 m/min and feed of 0.05 mm/rev. These parameters are applied in the digital process plan, ensuring consistent and standards-compliant machining recommendations.